\documentclass[11pt]{article}

\usepackage[margin=1in]{geometry}
\usepackage{amsmath,amssymb,amsfonts}
\usepackage{bm}
\usepackage{enumitem}
\usepackage{graphicx}
\usepackage{booktabs}
\usepackage{float}
\usepackage{placeins}
\usepackage{natbib}
\usepackage{hyperref}
\hypersetup{colorlinks=true, linkcolor=blue, citecolor=blue, urlcolor=blue}
\usepackage[english]{babel}
\usepackage{microtype}
\usepackage{setspace}
\usepackage{lineno}
\usepackage[title]{appendix}

\newcommand{\iid}{\stackrel{\text{iid}}{\sim}}
\DeclareMathOperator{\Pois}{Poisson}

\title{\bfseries Structured Bayesian Modeling of Dynamic Receptive Fields in Salamander Retinal Ganglion Cells}
\author{Alokesh Manna\\[2pt]
\normalsize\normalfont Department of Statistics, Texas A\&M University, College Station, TX, USA}
\date{}

\begin{document}

\maketitle

\begin{abstract}
\singlespacing
Neurons in the visual system are selective for specific spatial and temporal stimulus features, described by their \emph{receptive field}. Estimating one means a coefficient per pixel per time bin from few trials -- a high-dimensional problem requiring regularization. Sparse regularizers such as the LASSO handle the dimension but select pixels independently at each time point, with nothing to keep the region coherent in space or smooth in time; it can fragment or reorganize discontinuously even when the true response evolves smoothly, a failure since this evolving pattern is what a receptive-field estimate should capture. We formulate dynamic receptive-field estimation as a high-dimensional Bayesian problem: a Poisson model combining a Gaussian Markov random field in space with an autoregressive process in time, so the estimated field is smooth and coherent across space and time. On recordings from $155$ salamander retinal ganglion cells, fitting this model independently per neuron recovers a coherent surface, where a pixel-level Poisson-LASSO comparison instead returns a fragmented one. Summarizing each neuron's surface by its space-averaged temporal response and clustering these curves with a model-based functional-clustering procedure, BIC selects three balanced temporal-response phenotypes ($85$, $32$, $38$ neurons), against a degenerate grouping from clustering the raw surfaces. A simulation study with known ground truth confirms the same pattern, with the model beating an unregularized Poisson GLM, LASSO, and the elastic net on recovery and estimation accuracy, though LASSO controls false positives better. The per-neuron field identification, its contrast with LASSO, and the functional-clustering population typing constitute this paper's contribution.
\end{abstract}

\section{Introduction}
\label{sec:introduction}

Neurons in the visual system respond selectively to different spatial and temporal characteristics of visual stimuli. One important statistical object for characterizing this selectivity is the \emph{receptive field}, which describes the region of visual space and corresponding stimulus features that influence a neuron's activity. Estimating receptive fields from neural recordings is therefore an important problem in computational neuroscience.

As a motivating example, consider an experiment in which the firing activity of a retinal ganglion neuron from a salamander is recorded while a sequence of natural grayscale images is repeatedly presented as visual stimuli \citep{liu2022simple}; Section~\ref{sec:data} describes this dataset in full. For each image and each short time window after its onset, one observes a spike count, and the applied question is which pixels of the image, at which time lag, are associated with elevated firing. This is a large-scale localized-effect problem: the number of candidate spatio-temporal predictors is far larger than what a handful of trials per image can support, the predictors (neighboring pixels) are highly correlated, and any scientifically useful answer should point to a spatially coherent image region rather than an arbitrary subset of pixels.

Suppose that a visual stimulus is represented by a two-dimensional array of pixels. For each stimulus presentation, neural activity is recorded over a sequence of temporal bins. A natural statistical model is a Poisson regression in which the observed spike count is modeled as a function of the pixel intensities. Let $Y_{nkt}$ denote the spike count for neuron $k$, stimulus presentation $n$, and temporal bin $t$. We consider
\begin{equation}
Y_{nkt} \mid \lambda_{nkt} \sim \Pois(\lambda_{nkt}),
\end{equation}
with
\begin{equation}
\log(\lambda_{nkt}) = \alpha_k + \sum_{s=1}^{P} X_{ns}\beta_{kst},
\end{equation}
where $\alpha_k$ is neuron $k$'s intercept -- its baseline log firing rate when every pixel intensity is zero, i.e., the part of neuron $k$'s activity that the stimulus does not explain at all -- $X_{ns}$ denotes the intensity of pixel $s$ in the image shown on presentation $n$, $P$ is the number of pixels, and $\beta_{kst}$ represents the effect of pixel $s$ on neuron $k$'s firing at time $t$. We include $\alpha_k$ because neurons differ in overall excitability for reasons that have nothing to do with the stimulus (baseline metabolic state, electrode contact quality, and similar factors); without a neuron-specific intercept, this baseline difference would be absorbed into $\bm\beta_{kt}$ itself, inflating or deflating the fitted receptive-field coefficients by an amount that reflects nothing about stimulus selectivity. Fixing $\alpha_k$ as a separate term and estimating it jointly with $\bm\beta_{kt}$ removes this confound.

The collection
\begin{equation}
\bm{\beta}_{kt} = (\beta_{k1t},\ldots,\beta_{kPt})^\top
\end{equation}
can therefore be interpreted as a receptive-field surface. Importantly, this coefficient is not simply a generic regression parameter: its spatial organization has direct scientific meaning.

The dimensionality of the problem can become substantial. For example, a $13\times 13$ image observed over $30$ temporal bins gives
\begin{equation}
13\times13\times30=5070
\end{equation}
spatio-temporal coefficients for each neuron. When several neurons are observed, the number of parameters increases rapidly: with $155$ recorded neurons, a model giving every neuron its own free coefficient at every pixel and time bin, plus its own intercept $\alpha_k$, would have on the order of $155\times5{,}070\approx786{,}000$ parameters. This is precisely the regime in which standard Markov chain Monte Carlo (MCMC) becomes impractical: MCMC samples the parameter vector by updating one coordinate, or a small block of coordinates, at a time, and at this dimension -- with the strong spatial and temporal correlation a real receptive field has by construction -- such updates mix too slowly to reach convergence within a practical time budget. This is the concrete reason we turn to integrated nested Laplace approximation (INLA) rather than MCMC in Section~\ref{sec:model}: INLA replaces sampling the full high-dimensional coefficient field with numerical integration over a handful of hyperparameters, exploiting exactly the Gaussian structure this paper places on $\bm\beta_{kt}$ (Section~\ref{sec:inla-inference} states this precisely).

A common solution to high-dimensional regression is the LASSO, which estimates coefficients using an $L_1$ penalty. Although this produces sparse models, ordinary LASSO does not explicitly encode the spatial organization of neighboring pixels. This is potentially problematic in receptive-field estimation because neighboring pixels are often strongly related, and biologically meaningful receptive fields are generally expected to occupy spatially organized regions rather than consist of unrelated isolated pixels.

The limitations of LASSO in the presence of dependent predictors have been studied extensively. \citet{zou2005} showed that LASSO can behave poorly when predictors are strongly correlated and introduced the elastic net partly to obtain a grouping effect among correlated predictors. More broadly, \citet{freijeirogonzalez2022} reviewed variable-selection behavior under dependence among covariates and emphasized that the assumptions required for reliable LASSO selection can be restrictive in dependent high-dimensional settings.

These issues motivate a different formulation for receptive-field estimation. Instead of treating every pixel as an independent candidate variable, we propose to model the receptive field as a structured spatio-temporal object and to borrow information across neurons belonging to the same functional type.

The central objective of this work is therefore to develop a hierarchical Bayesian model that simultaneously addresses:
\begin{enumerate}[label=(\arabic*)]
\item spatial dependence among neighboring pixels;
\item temporal dependence in receptive-field effects;
\item sparsity of the relevant visual region;
\item heterogeneity across neurons; and
\item shared structure across neuron types.
\end{enumerate}

This leads to the following scientific questions:
\begin{itemize}[leftmargin=2em]
\item Which spatial regions of the stimulus influence neural firing?
\item How does the receptive field evolve over time?
\item Which features of receptive fields are shared by neurons of the same type?
\item How much do individual neurons deviate from their type-level receptive field?
\item Can structured sparsity recover biologically meaningful receptive-field regions more accurately and stably than conventional LASSO?
\end{itemize}

We separate what this paper demonstrates from what it targets. The spatially and temporally structured single-neuron model of Section~\ref{sec:model} -- Gaussian Markov random field in space, autoregressive process in time -- is fit, independently for each of 155 neurons, throughout Sections~\ref{sec:real-data}--\ref{sec:population-typing}, and its fitted surfaces are used to recover a population-level typing of the neurons. The shared type-level component $\mu_{g(k),st}$ and structured-sparsity prior that complete the fully hierarchical specification of Section~\ref{sec:model} are the target this per-neuron model is designed to extend to; they are specified in full but not yet fit jointly across the population, for reasons and with consequences discussed explicitly in Section~\ref{sec:limitations}.

Stated plainly: none of SPDE-based spatial priors, AR(1) temporal processes, or INLA is individually new -- each is established methodology, cited throughout. The statistical contribution is formulating dynamic receptive-field estimation as a single structured latent-Gaussian model that combines all three, together with the type-sharing and structured-sparsity extension of Section~\ref{sec:future-work}, into one coherent representation of the problem stated in Section~\ref{sec:motivation} -- spatial coherence, temporal evolution, neuron-specific heterogeneity, and shared structure across types simultaneously -- and showing that this structured representation is what makes fitting tractable at this dataset's scale ($5{,}070$ pixel-by-time coefficients per neuron, $155$ neurons) -- a scale at which a generic treatment of the same latent field, one that did not exploit this structure and instead relied on standard MCMC, would not be tractable within a practical time budget (Section~\ref{sec:inla-inference}).

This work also connects to several related methodological threads, on both the spatial and the temporal side of the model. On the spatial side, \citet{manna2026bwmp} develop a graph-Laplacian-based operator construction for large-scale multivariate spatial fields, in a similar computational spirit to the SPDE representation adopted here, though built from a different operator and applied to spatial transcriptomics data rather than neural receptive fields; \citet{manna2026scalablespatialpointprocess} apply INLA-based latent Gaussian spatial point process models to a differently structured large-scale spatial inference problem (forensic footwear evidence), sharing this paper's reliance on the same INLA hyperparameter-dimension scaling argument used in Section~\ref{sec:inla-inference}. On the temporal side, \citet{manna2025arlagselection} develop a joint Bayesian selection mechanism over predictors and autoregressive lag order for environmental and financial time series, related in spirit -- though again in a different application domain -- to the AR(1) structure placed on $\mathbf{w}_{kt}$ in Section~\ref{sec:model}. More broadly, statistical modeling of neural spike-train data has a substantial independent literature beyond the Poisson-GLM receptive-field work of \citet{pillow2008spatio} already cited in Section~\ref{sec:data}, including general probability models for single and multiple spike trains \citep{kassventura2001,paninski2008statistical} and spatio-temporal point-process approaches to event data exhibiting clustering or heavy-tailed, extreme behavior \citep{reinhart2018review,davismikosch2008}. These point-process approaches frame a closely related class of problems -- localizing and characterizing structured effects in space and time -- through an event-intensity lens, in contrast to the pixel-coefficient regression lens adopted throughout this paper; we note the connection rather than adopt the framework, since our target is a smooth, spatially coherent coefficient surface rather than an event-rate intensity per se. The spatial, temporal, and spatio-temporal methodological threads sketched above -- together with their extension to neuroscience applications of the kind studied here -- are developed at greater length in \citet{manna2026dissertation}.

\section{Neural Spike Data}
\label{sec:data}

We consider a neuroscience experiment that recorded the firing activity of a retinal ganglion neuron from a salamander \citep{liu2022simple}. This follows a substantial body of work using Poisson generalized linear models to characterize the spatio-temporal receptive fields of retinal ganglion cell populations from spike-train data \citep{pillow2008spatio}. The experiment involved repeatedly presenting 303 distinct grayscale images, with each image shown 13 times, resulting in 3{,}939 trials. Each trial lasted 200 milliseconds, during which the number of neuronal spikes was counted; we divided this 200-millisecond span into 30 equally spaced time intervals and, for each trial, counted the number of neural firings within each interval. The images were originally $256\times256$ pixels, which we coarsened down to $13\times13$. We chose one neuron and sought to understand which particular spatial locations in the $13\times13$ images are likely to spark its firing. This dataset is well-suited to the modeling framework developed below, as it contains rich spatial structure from the image stimuli, temporal variability across trials, and repeated measures that enable us to disentangle stimulus-driven effects from intrinsic neuronal noise.

\begin{figure}[H]
    \centering
    \includegraphics[width=0.8\linewidth]{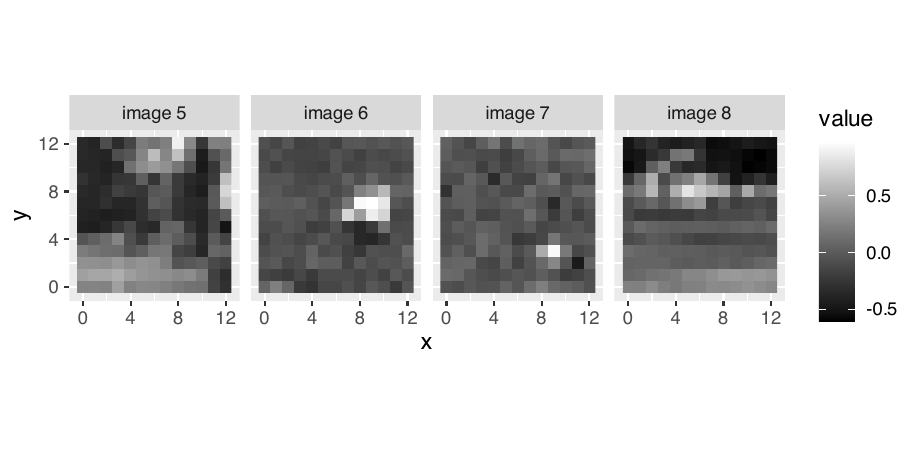}
    \caption{Four representative $13\times 13$ stimulus images (mean-centered pixel intensity) from the salamander retinal ganglion cell dataset.}
    \label{fig:image_example}
\end{figure}

Figure~\ref{fig:image_example} shows four representative stimulus images after coarsening to the $13\times13$ pixel grid used throughout this paper. Pixel intensity is mean-centered per image (grayscale value minus the image's own mean), which is the covariate scale used in every model fit that follows.

\section{Motivation: Why LASSO Is Not Sufficient}
\label{sec:motivation}

\subsection{Ordinary sparse regression}

The LASSO estimates a coefficient vector by solving an optimization problem of the form
\begin{equation}
\widehat{\bm{\beta}} = \arg\min_{\bm{\beta}} \left\{ -\ell(\bm{\beta}) + \lambda\sum_{s=1}^{P}|\beta_s| \right\},
\end{equation}
where $\ell(\bm{\beta})$ is the log-likelihood and $\lambda\ge0$ is the penalty parameter controlling the amount of shrinkage -- a different quantity from the Poisson rate $\lambda_{nkt}$ of Section~\ref{sec:introduction}, a standard notational overload in this literature that we flag once here rather than let the shared letter cause confusion later.

The $L_1$ penalty encourages many coefficients to be exactly zero, which is attractive for high-dimensional receptive-field estimation.

However, it does not distinguish between two very different configurations:
\begin{equation}
\begin{pmatrix}
0&0&0&0\\
0&1&1&0\\
0&1&1&0\\
0&0&0&0
\end{pmatrix}
\qquad \text{and} \qquad
\begin{pmatrix}
1&0&0&0\\
0&0&1&0\\
0&0&0&0\\
0&1&0&1
\end{pmatrix}.
\end{equation}
Both configurations contain exactly four nonzero coefficients, but the first represents a spatially coherent receptive field whereas the second is spatially fragmented, with no two active pixels adjacent to one another. The $L_1$ penalty itself does not encode this distinction.

\subsection{A motivating empirical illustration}
\label{sec:empirical-illustration}

The fragmentation problem above is not merely hypothetical: using the neural spike data introduced in Section~\ref{sec:data}, we now show that it occurs in practice, and that a coarser, image-level analysis cannot resolve it either.

\begin{figure}[H]
    \centering
    \includegraphics[width=0.8\linewidth]{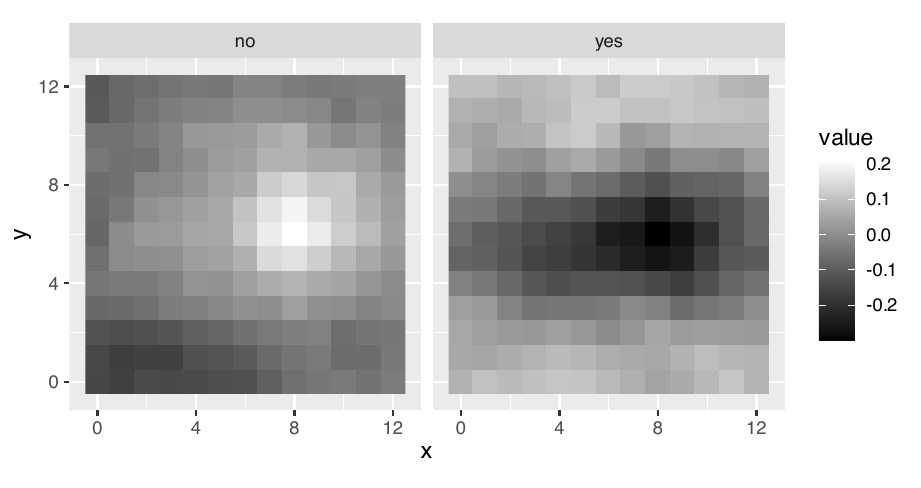}
    \caption{Pixel-wise average stimulus image for images classified as more strongly firing-associated (``yes'') versus less firing-associated (``no'') under the image-level Poisson baseline described above.}
    \label{fig:yes_no_spike}
\end{figure}

We first ask a deliberately coarse diagnostic question: does the \emph{whole image}, rather than any particular pixel, predict elevated firing? For image $n\in\{1,\ldots,303\}$, let $\bar y_n$ denote the mean spike count across all trials and time bins in which that image was shown; we split the images into a ``yes'' group (the more strongly firing-associated half, ranked by $\log\bar y_n$) and a ``no'' group, and average the (mean-centered) pixel intensities within each. Figure~\ref{fig:yes_no_spike} shows the result: the ``yes'' group has a visibly darker horizontal band through the middle rows relative to the ``no'' group's small, localized bright patch — an early hint that stimulus energy in that mid-image band is associated with elevated firing. Because this comparison operates at the level of whole images, however, it says nothing about \emph{where}: it cannot by itself localize a receptive field within the $13\times13$ grid.

\begin{figure}[H]
    \centering
    \includegraphics[width=0.6\linewidth]{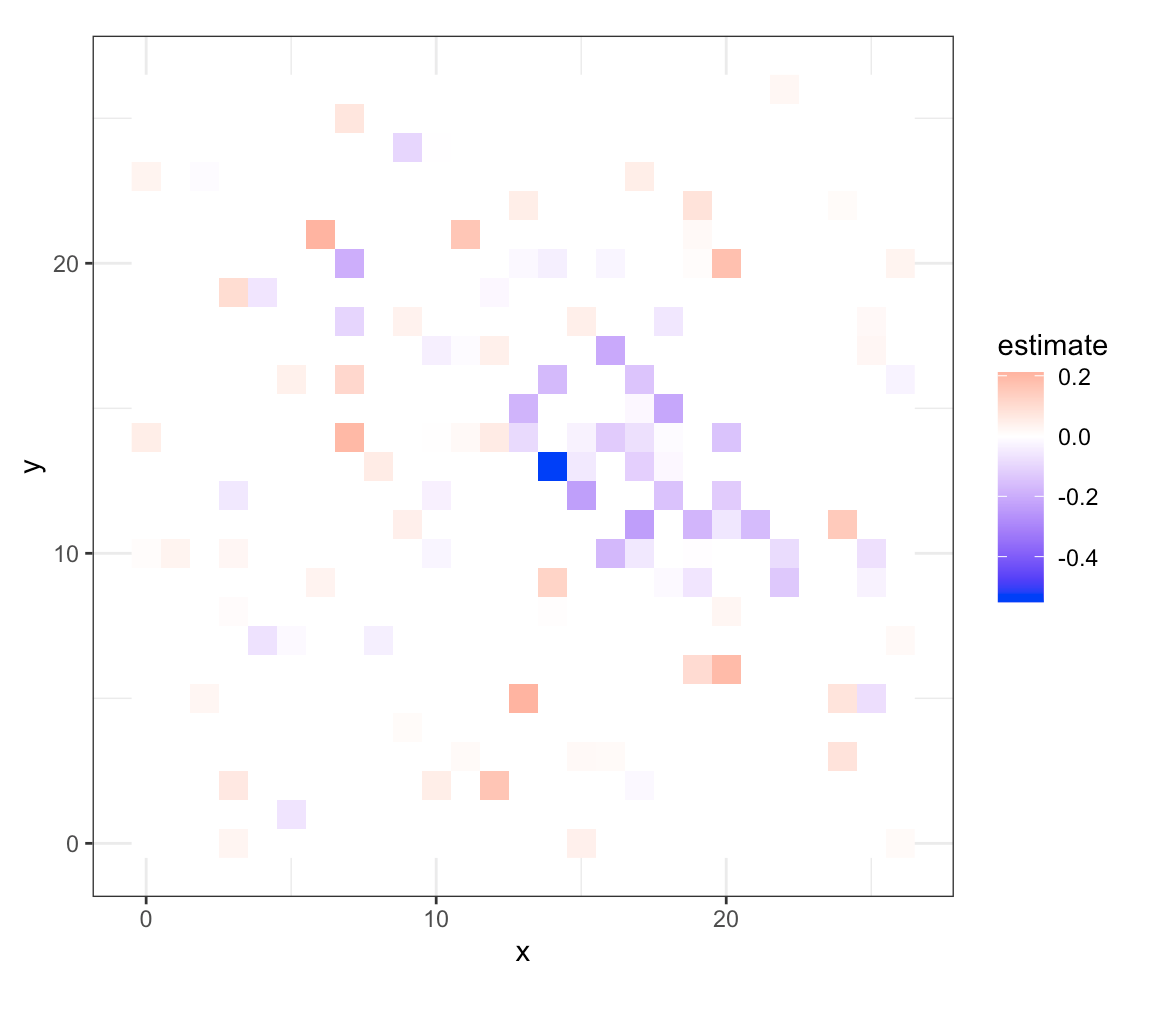}
    \caption{Pixel-level Poisson-LASSO fit at a fixed illustrative penalty ($\lambda=0.02$), 27 of 169 pixel coefficients nonzero. Beyond the contiguous central cluster, several isolated, spatially disconnected pixels are also selected.}
    \label{fig:lasso_fit}
\end{figure}

To localize the effect, we next fit an ordinary pixel-level Poisson-LASSO regression, pooled across all 30 time bins and 13 repeats, with no held-out split and no spatial or temporal smoothing. Writing $y_n$ for the spike count at observation $n$ (one of $30\times13\times303=118{,}170$ time-bin/repeat/image combinations), $X_{ns}$ for pixel $s$'s mean-centered intensity in the corresponding image, and $\mathbf{x}_n = (X_{n1},\ldots,X_{n,169})^\top$ for the vector collecting all $169$ pixel intensities of that observation,
\begin{equation}
\widehat{\bm{\beta}}_{\mathrm{lasso}} = \arg\min_{\bm{\beta}}\left\{-\sum_n \left(y_n\, \mathbf{x}_n^\top\bm{\beta} - \exp(\mathbf{x}_n^\top\bm{\beta})\right) + \lambda\sum_{s=1}^{169}|\beta_s|\right\},
\end{equation}
where $\lambda\ge0$ is the same LASSO penalty parameter as in Section~\ref{sec:motivation} above. This illustrative fit uses a single fixed penalty ($\lambda=0.02$) on the full pooled data, rather than a cross-validated one, and gives a sparse fit with 27 of the 169 pixel coefficients nonzero. Figure~\ref{fig:lasso_fit} shows this fit. The LASSO does recover a small contiguous block of negative coefficients — consistent with the mid-image band flagged by the image-level baseline above — but it also selects several isolated, spatially disconnected pixels scattered near the image border that share no boundary with the main cluster or with one another. The $L_1$ penalty has no mechanism to prefer a spatially coherent active set over a scattered one with the same number of nonzero entries, so isolated-pixel selections with no clear biological interpretation and a genuine receptive-field cluster are penalized identically.

Together, Figures~\ref{fig:yes_no_spike} and \ref{fig:lasso_fit} motivate the remainder of the paper: Section~\ref{sec:model} develops a hierarchical Bayesian model whose spatial prior encourages neighboring pixels to be similar, which discourages -- though does not by itself guarantee -- the kind of isolated, disconnected selections seen in Figure~\ref{fig:lasso_fit} (spatial smoothness and a formally sparse, connected support are distinct properties, and the discussion in Section~\ref{sec:model} returns to this distinction). Section~\ref{sec:real-data} applies the smoothed model to this same neuron, completing the comparison with an explicit spatio-temporal fit.

\subsection{Correlated image predictors}

Neighboring pixels in an image are generally correlated. If two nearby pixels contain similar information, it may be difficult to determine which individual pixel should receive the nonzero coefficient.

This is a well-known issue in LASSO variable selection under correlated predictors. \citet{zou2005} specifically proposed the elastic net because LASSO may select only one variable from a group of strongly correlated variables, whereas the elastic net encourages correlated variables to enter or leave the model together.

The problem is particularly relevant here because the scientific target is not necessarily an individual pixel. Rather, the target is a spatial region representing a receptive field.

\subsection{Shrinkage bias}

The LASSO also shrinks all nonzero coefficients toward zero. Consequently, sparsity and estimation are obtained simultaneously through the same penalty.

For receptive-field analysis this can be undesirable because the magnitude of $\beta_{kst}$ can itself be scientifically meaningful. A method that strongly shrinks large effects may distort the estimated strength of neuronal responses.

\subsection{Spatial and temporal dependence}

The receptive field is not expected to change arbitrarily from one pixel to another or from one temporal bin to the next. We therefore expect $\beta_{k,s,t}$ to be related to $\beta_{k,s',t}$ when $s$ and $s'$ are neighboring pixels, and to $\beta_{k,s,t-1}$ when adjacent temporal bins are considered. Ordinary LASSO does not explicitly incorporate either relationship. This is not merely a modeling convenience: \citet{park2011receptive} show that imposing an explicit spatial-locality prior on receptive-field coefficients yields substantially more accurate and interpretable estimates than pixelwise-independent regularization, particularly when data are limited relative to the number of stimulus dimensions.

\subsection{Heterogeneous neurons}

Finally, a collection of neurons may contain several functional types. Two neurons of the same type may have similar receptive-field organization, while neurons belonging to different types may have systematically different spatial or temporal response patterns.

Treating every neuron independently ignores this potentially valuable source of information. Conversely, forcing all neurons to have the same receptive field ignores biological heterogeneity. This motivates a hierarchical model that allows both sharing and differentiation.

\section{The Hierarchical Structured-Sparsity Model}
\label{sec:model}

\paragraph{What we fit, stated first.} Every result in this paper comes from one model: a Bayesian hierarchical model, fit independently for each of the $155$ recorded neurons, that treats a neuron's receptive-field surface $\bm\beta_{kt}$ as a latent Gaussian object with (i) a neuron-specific intercept $\alpha_k$ absorbing baseline excitability, (ii) an i.i.d.\ random effect for repeated presentations of the same image, (iii) a spatially structured Gaussian Markov random field (via an SPDE representation) that ties the coefficient at each pixel to its neighbors, and (iv) an AR(1) process linking that spatial field smoothly across time bins. ``Hierarchical'' here refers to this model's own structure -- a top level of a small number of hyperparameters governing a lower-level latent Gaussian field, in turn governing the observed spike counts -- fit by INLA for the reasons given in Section~\ref{sec:inla-inference}, not to pooling across neurons: this model is fit to each neuron on its own, with no information shared between neurons. Section~\ref{sec:model-fitted} (Part A) states this model precisely; it is the only model whose results appear anywhere in Sections~\ref{sec:real-data}--\ref{sec:population-typing} or the simulation study of Section~\ref{sec:simulation}. Section~\ref{sec:future-work} (Part B) briefly sketches, without presenting as a result, one direction this per-neuron model could be extended to pool information across neurons of the same functional type; that extension has not been implemented or fit, and no number reported anywhere in this paper depends on it.

\subsection{Neural response model and observation family}

Let $Y_{nkt}$ denote the spike count of neuron $k$ during temporal bin $t$ for stimulus presentation $n$. Both parts below share the same observation model,
\begin{equation}
Y_{nkt} \mid \lambda_{nkt} \sim \Pois(\lambda_{nkt}), \qquad \log \lambda_{nkt} = \alpha_k + \mathbf{X}_n^\top\bm{\beta}_{kt},
\end{equation}
where $\alpha_k$ is neuron $k$'s intercept, as defined in Section~\ref{sec:introduction} above, $\mathbf{X}_n = (X_{n1},\ldots,X_{nP})^\top$ contains the pixel intensities and $\bm{\beta}_{kt} = (\beta_{k1t},\ldots,\beta_{kPt})^\top$ is the receptive-field coefficient vector for neuron $k$ at time $t$; the sign and magnitude of $\beta_{kst}$ describe the association between stimulus intensity at spatial location $s$ and neural firing at time $t$. What the two parts differ on is how $\bm{\beta}_{kt}$ is modeled -- independently per neuron (Part A) or through a shared type-level component (Part B) -- not the observation family itself.

\paragraph{Why Poisson.} Spike counts $Y_{nkt}$ are non-negative integers recorded over a fixed time window, which is precisely the sampling situation the Poisson distribution is built for, and it is the standard observation model for spike-train data in this literature \citep{pillow2008spatio}. The log link guarantees $\lambda_{nkt}>0$ for any real-valued $\bm\beta_{kt}$, which is what lets us place an unconstrained Gaussian (GMRF/SPDE) prior directly on the coefficient surface in Part A below without a separate positivity constraint. The Poisson model does impose the restriction that the conditional variance equals the conditional mean; we have not formally tested this assumption against the fitted neuron's residuals, and do not claim it holds, which is why the negative-binomial extension is listed as explicit, not yet completed, follow-up work in Section~\ref{sec:limitations} rather than silently assumed away. A dispersion check (e.g., comparing the Pearson $\chi^2$ statistic to its degrees of freedom, or refitting with a negative-binomial likelihood and comparing by WAIC) is a low-cost addition we flag for the revision that completes Section~\ref{sec:limitations}.

\subsection{Part A --- the model we fit}
\label{sec:model-fitted}

For each neuron $k$, independently, we model the coefficient surface $\bm\beta_{kt}$ directly -- there is no type index $g$ anywhere in this subsection -- as a spatially smoothed, temporally autoregressive Gaussian field, fit by INLA; every quantity below carries the neuron index $k$ to keep this explicit. Concretely, following the SPDE approach of \citet{bakka2018spatial}, we represent the spatial component for neuron $k$ as a continuous-domain Gaussian random field $w_k(\mathbf{u})$ over the pixel grid, discretized on a triangulated mesh (shared across neurons) with basis functions $\psi_1,\ldots,\psi_M$,
\begin{equation}
w_k(\mathbf{u}) = \sum_{m=1}^{M} \psi_m(\mathbf{u})\, w_{km}, \qquad \mathbf{w}_k = (w_{k1},\ldots,w_{kM})^\top,
\end{equation}
where $M$ is the number of mesh nodes. Writing $\mathbf{A}\in\mathbb{R}^{P\times M}$ for the projector matrix, shared across neurons, that evaluates the mesh-based field at the $P=169$ pixel centers, the pixel-level spatial coefficient for neuron $k$ at time group $t$ is $\bm\beta^{\mathrm{sp}}_{kt} = \mathbf{A}\mathbf{w}_{kt}$, and the mesh-node field evolves across the $T=30$ time groups by an AR(1) process,
\begin{equation}
\mathbf{w}_{kt} = \rho_k\,\mathbf{w}_{k,t-1} + \bm\epsilon_{kt}, \qquad \bm\epsilon_{kt} \iid N(\mathbf{0}, Q_w^{-1}), \qquad |\rho_k|<1, \qquad t=2,\ldots,T,
\end{equation}
with $\mathbf{w}_{k1} \sim N\!\left(\mathbf{0},\, \tfrac{1}{1-\rho_k^2}Q_w^{-1}\right)$ the stationary marginal distribution implied by this recursion -- matching the group-AR(1) construction actually fit by INLA below, rather than left unstated -- and a penalized-complexity prior on $\rho_k$.

\paragraph{How $Q_w$ governs spatial smoothing.} $Q_w$ is not an arbitrary positive-definite matrix chosen for convenience: it is built from a finite-element discretization of a differential operator on the mesh \citep{lindgren2011explicit}, sparse by construction, with a zero entry for every pair of mesh nodes that are not spatial neighbors -- the same graph/mesh-adjacency precision-matrix mechanism as other point-process literature on forensic footwear-impression intensity surfaces \citep{manna2026scalablespatialpointprocess}. This sparsity pattern is what defines the smoothing: by the Gaussian Markov random field conditional-mean identity, each mesh-node coefficient is pulled only toward a precision-weighted average of its immediate neighbors, never toward non-adjacent nodes, which is the precise mechanism discouraging the isolated, disconnected pixel selections of Figure~\ref{fig:lasso_fit}. Appendix~\ref{app:precision-detail} gives the full construction of $Q_w$ and works out exactly how its two hyperparameters $(\kappa,\sigma_w)$ separately control the range and strength of this coupling versus its overall scale.

The full per-neuron model fit in Sections~\ref{sec:real-data}--\ref{sec:population-typing} is then
\begin{equation}
\log\lambda_{nkt} = \alpha_k + b_{r(n)} + \left(\mathbf{X}_n^\top\mathbf{A}\right)\mathbf{w}_{kt}, \qquad b_{r(n)}\sim N(0,1/\tau_b),
\end{equation}
where $\alpha_k$ is neuron $k$'s fixed intercept -- the same symbol introduced in the shared observation model of Section~\ref{sec:model} above, kept consistent here rather than renamed -- $r(n)\in\{1,\ldots,13\}$ indexes which of the $13$ repeated presentations observation $n$'s trial belongs to -- the same $13$ repetition labels are shared across all $303$ images, so $r(n)$ is not a unique per-trial identifier -- $b_{r(n)}$ is an i.i.d.\ Gaussian random effect absorbing variability associated with repetition number at fixed image and time bin, and $\mathbf{w}_{kt}$ is neuron $k$'s own mesh-node field at time group $t$, fit independently of every other neuron: there is no pooling of $\mathbf{w}_{kt}$ across $k$ in Part A. Because $Q_w$, $\rho_k$, and $\tau_b$ are a small number of hyperparameters conditional on which the whole latent field $(\alpha_k, \mathbf{b}, \mathbf{w}_{k1},\ldots,\mathbf{w}_{kT})$ is jointly Gaussian, this is a latent Gaussian model; Section~\ref{sec:inla-inference} below states this precisely -- the parameter vector, the priors, the joint posterior, and the formal LGM representation -- and we fit it by INLA rather than MCMC for the reasons given there. Equivalently, writing $\beta_{ijt} = (\mathbf{A}\mathbf{w}_{kt})_{ij}$ for the pixel-level coefficient this induces for neuron $k$ at pixel $(i,j)$ (the neuron index is dropped from $\beta_{ijt}$ below, as Section~\ref{sec:real-data} fixes a single neuron throughout), the model can be read as the simpler-looking
\begin{equation}
\log\lambda_{nt} = \alpha + \sum_{i=1}^{13}\sum_{j=1}^{13}\beta_{ijt}X_{nij} + b_{r(n)}
\end{equation}
used to introduce the fit in Section~\ref{sec:real-data}, with the understanding that $\beta_{ijt}$ is not a free parameter per pixel and time bin but is generated by projecting the much lower-dimensional mesh field $\mathbf{w}_t$ onto the pixel grid -- this is exactly what makes the $5{,}070$-coefficient surface tractable to estimate from $3{,}939$ trials in the first place -- and the neuron index is dropped from $\alpha_k$ for the same reason it is dropped from $\beta_{ijt}$.

\subsection{Bayesian inference and the role of INLA}
\label{sec:inla-inference}

We now make precise, for the model just stated, exactly what is being inferred and why INLA rather than Markov chain Monte Carlo (MCMC) is used to fit it -- the same question Section~\ref{sec:real-data} raises informally for the real-data fit is answered here once, with notation, for Part A in general.

\paragraph{Parameter vector.} Fix a neuron $k$ and, following it, drop the subscript $k$ from every quantity below except where it is needed for clarity, exactly as Section~\ref{sec:model-fitted} already does for $\beta_{ijt}$ and $\alpha_k$. Collect the latent Gaussian quantities of the model into a single vector,
\begin{equation}
\bm\theta := \big(\alpha,\ \mathbf{b},\ \mathbf{w}_1,\ldots,\mathbf{w}_T\big) \in \mathbb{R}^d, \qquad d = 1 + 13 + MT,
\end{equation}
where $\mathbf{b} = (b_1,\ldots,b_{13})^\top$ collects the repetition random effects and $\mathbf{w}_t\in\mathbb{R}^M$ is the mesh-node field at time group $t$ from Section~\ref{sec:model-fitted}. Separately, collect the hyperparameters that govern the distribution of $\bm\theta$ into
\begin{equation}
\bm\psi := (\kappa,\ \sigma_w,\ \rho,\ \tau_b),
\end{equation}
where $(\kappa,\sigma_w)$ are the SPDE range and marginal-variance hyperparameters that determine the spatial precision $Q_w = Q_w(\kappa,\sigma_w)$ of Section~\ref{sec:model-fitted} \citep{lindgren2011explicit,bakka2018spatial}, $\rho$ is the AR(1) persistence of $\mathbf{w}_t$ across time groups, and $\tau_b$ is the precision of the repetition random effect. Unlike the fixed intercept $\alpha$, which we give a fixed-variance Gaussian prior with no hyperparameter of its own, every other block of $\bm\theta$ has its distribution governed by $\bm\psi$.

\paragraph{Likelihood.} With $n=1,\ldots,3{,}939$ trials indexing $(\text{image}, \text{time bin}, \text{repetition})$ triples for the fixed neuron, the observation model of Section~\ref{sec:model} gives
\begin{equation}
\prod_{n=1}^{3939} P(Y_n = y_n \mid \bm\theta) = \prod_{n=1}^{3939} \frac{e^{-\lambda_n}\lambda_n^{y_n}}{y_n!}, \qquad \log\lambda_n = \eta_n(\bm\theta) = \alpha + b_{r(n)} + \left(\mathbf{X}_n^\top\mathbf{A}\right)\mathbf{w}_{t(n)}.
\end{equation}
Note that $\eta_n(\bm\theta)$ is linear in $\bm\theta$: writing $\mathbf{b}_n\in\mathbb{R}^d$ for the vector with a $1$ in $\bm\theta$'s $\alpha$-coordinate, a $1$ in its $b_{r(n)}$-coordinate, $(\mathbf{X}_n^\top\mathbf{A})$ in its $\mathbf{w}_{t(n)}$-block, and zero everywhere else, $\eta_n(\bm\theta) = \mathbf{b}_n^\top\bm\theta$ exactly -- the same linear-predictor structure used throughout the INLA framework below.

\paragraph{Priors.} We place independent priors on each block of $\bm\theta$ given $\bm\psi$,
\begin{equation}
\alpha \sim N(0, 1000), \qquad b_r \mid \tau_b \iid N(0,\tau_b^{-1}) \text{ for } r=1,\ldots,13,
\end{equation}
\begin{align}
\mathbf{w}_1 \mid \kappa,\sigma_w &\sim N\!\left(\mathbf{0}, \tfrac{1}{1-\rho^2}Q_w(\kappa,\sigma_w)^{-1}\right), \\
\mathbf{w}_t \mid \mathbf{w}_{t-1},\bm\psi &\sim N\!\left(\rho\,\mathbf{w}_{t-1},\, Q_w(\kappa,\sigma_w)^{-1}\right), \qquad t=2,\ldots,T,
\end{align}
matching Section~\ref{sec:model-fitted}'s recursion, and penalized-complexity hyperpriors on $\bm\psi$ \citep{bakka2018spatial}: $(\kappa,\sigma_w)$ reparametrized to a spatial range and marginal standard deviation with $P(\text{range}<r_0)=0.05$ and $P(\sigma_w>s_0)=0.05$ for data-scale-informed $(r_0,s_0)$, a penalized-complexity prior on $\rho\in(-1,1)$ shrinking toward $\rho=0$, and $\tau_b \sim \text{Exponential}(\lambda_b)$. Conditional on $\bm\psi$, every block of $\bm\theta$ is Gaussian, and the blocks are mutually independent given $\bm\psi$, so
\begin{equation}
\bm\theta \mid \bm\psi \sim N\!\left(\mathbf{0},\, Q(\bm\psi)^{-1}\right), \qquad Q(\bm\psi) = \mathrm{blkdiag}\!\left(1000^{-1},\ \tau_b\mathbf{I}_{13},\ Q_w(\kappa,\sigma_w,\rho)\right),
\end{equation}
where $Q_w(\kappa,\sigma_w,\rho)\in\mathbb{R}^{MT\times MT}$ is the block-tridiagonal precision induced by the AR(1)-over-SPDE recursion above -- the standard ``group model'' construction of \citet{bakka2018spatial} -- rather than written out entry by entry here.

\paragraph{Joint posterior.} Combining the likelihood and priors, the target of inference is
\begin{equation}
\pi(\bm\theta,\bm\psi \mid \mathbf{y}) \ \propto\ \left[\prod_{n=1}^{3939} \frac{e^{-\lambda_n(\bm\theta)}\lambda_n(\bm\theta)^{y_n}}{y_n!}\right] \times \sqrt{\frac{|Q(\bm\psi)|_*}{(2\pi)^d}}\exp\!\left(-\tfrac{1}{2}\bm\theta^\top Q(\bm\psi)\bm\theta\right) \times \pi(\bm\psi),
\end{equation}
with $d=1+13+MT$ as above. As in Section~\ref{sec:real-data}, $d$ is large enough (on the order of the $5{,}070$ pixel-level coefficients this field projects to) that reliable MCMC over $\bm\theta$ jointly with $\bm\psi$ is not computationally feasible within a reasonable time frame at this scale; we instead exploit the structure of the joint posterior above directly, via the LGM representation given next.

\paragraph{Framing as a latent Gaussian model.} A latent Gaussian model (LGM) is any hierarchical model of the form
\begin{align}
\bm\psi &\sim \pi(\bm\psi), \qquad \bm\theta \mid \bm\psi \sim N\!\left(\mathbf{0}, Q(\bm\psi)^{-1}\right), \\
y_n \mid \bm\theta,\bm\psi &\iid p(y_n\mid \eta_n(\bm\theta),\bm\psi), \qquad \eta_n(\bm\theta) = \mathbf{b}_n^\top\bm\theta \text{ for some } \mathbf{b}_n\in\mathbb{R}^d,
\end{align}
i.e., a Gaussian latent field entering the likelihood only through linear predictors $\eta_n(\bm\theta)$ \citep{rue2009approximate,rue2017Bayesian}. The model above satisfies this exactly: $\bm\theta\mid\bm\psi$ is Gaussian by the Priors paragraph, and $\eta_n(\bm\theta)=\mathbf{b}_n^\top\bm\theta$ is linear in $\bm\theta$ by the Likelihood paragraph, with no additional likelihood hyperparameter (the Poisson family has none).

\paragraph{Why scalability is a real obstacle here, concretely.} A generic MCMC sampler for $\pi(\bm\theta,\bm\psi\mid\mathbf{y})$ would update, at every iteration, a $d$-dimensional vector with $d=1+13+MT$ -- on the order of the $5{,}070$ pixel-level coefficients this field projects to (Section~\ref{sec:data}) -- and the components of $\bm\theta$ are highly correlated with one another by construction, since that correlation is exactly what the SPDE/AR(1) prior encodes. Two things follow directly. First, a sampler that updates coordinates one at a time (Gibbs) or in small blocks mixes slowly precisely because of this correlation: strongly correlated neighbors mean each individual update moves the chain only a small distance in the directions that matter, so the number of iterations needed to explore the posterior grows with both $d$ and the strength of the spatial/temporal dependence, not with $d$ alone. Second, even a well-tuned joint sampler (e.g., Hamiltonian Monte Carlo) still requires repeated likelihood and gradient evaluations over the full $d$-dimensional state at every iteration, and assessing convergence reliably at this scale -- multiple chains, thousands of iterations each, checked for a $d$-dimensional target -- is itself a substantial computational commitment. None of this is infeasible in principle; it is infeasible \emph{within a reasonable time frame at the scale of this paper's data} ($n=3{,}939$ trials per neuron, $155$ neurons), which is the standard we hold ourselves to throughout.

\paragraph{How INLA avoids this.} INLA does not sample $\bm\theta$ at all. Because $\bm\theta\mid\bm\psi$ is exactly Gaussian (the LGM property above) and the Poisson likelihood is log-concave, the conditional posterior $\pi(\bm\theta\mid\bm\psi,\mathbf{y})$ is extremely well approximated by a Gaussian via a Laplace approximation, computed by direct numerical optimization (not sampling) exploiting the sparsity of $Q(\bm\psi)$ -- the SPDE/AR(1) precision matrix is sparse by construction, since each pixel or time step is directly coupled only to its neighbors, which is exactly what makes this optimization fast even for $d$ in the thousands. The only integration that remains is over $\bm\psi$ itself -- four scalars here -- carried out by evaluating this Laplace approximation on a small grid or a nested-approximation scheme in $\bm\psi$-space. So the computational cost that would otherwise grow with $d$ (thousands) is replaced by a cost that grows with $\dim(\bm\psi)=4$: integrated nested Laplace approximation \citep[INLA;][]{rue2009approximate,rue2017Bayesian} approximates $\pi(\bm\psi\mid\mathbf{y})$ and each marginal $\pi(\theta_j\mid\mathbf{y})$ this way. This is the same computational-advantage argument we have used, on a differently structured but similarly large-dimensional latent Gaussian field, in \citet{manna2026scalablespatialpointprocess}: INLA's cost is driven by the dimension of $\bm\psi$, not of $\bm\theta$, which is what makes fields of this size tractable at all, without the mixing and convergence-assessment burden described above. We fit $\bm\theta,\bm\psi$ by INLA throughout the real-data analysis (Section~\ref{app:real-data}) on this basis, using the SPDE representation of $Q_w$ and the AR(1) group model as implemented in \texttt{R-INLA} \citep{bakka2018spatial}.

Part A above is the whole of the model fit anywhere in this paper. A target hierarchical extension -- pooling information across neurons of the same functional type, with a sparsity mechanism on top of the smoothing above -- is sketched briefly, as a direction rather than a fully specified model, in Section~\ref{sec:future-work} (Future Work); it has not been fit, and no result anywhere in this paper depends on it. We state that separation once, here, rather than interleave a not-yet-fit model with the one whose results follow.

\section{Real-Data Analysis}
\label{sec:real-data}
\label{app:real-data}

This section reports the real-data analysis in two parts: a single-neuron demonstration of the fitted model (Section~\ref{sec:single-neuron-demo}), and a population-level typing of all $155$ recorded neurons built from the same per-neuron fit (Section~\ref{sec:population-typing}). It is real data throughout, with a real, but not known, receptive field -- see Section~\ref{sec:simulation} for why that distinction matters and for the corresponding simulation comparison, which is the paper's primary quantitative evidence for how the proposed model compares against simpler non-hierarchical alternatives, evaluated there under known ground truth.

\subsection{Single-neuron demonstration}
\label{sec:single-neuron-demo}

Section~\ref{sec:empirical-illustration} showed, on the neural spike data of Section~\ref{sec:data}, that an image-level Poisson baseline cannot localize the receptive field (Figure~\ref{fig:yes_no_spike}) and that an ordinary pixel-level Poisson-LASSO fit, while it does localize, returns a spatially fragmented coefficient pattern (Figure~\ref{fig:lasso_fit}). We now complete that comparison by fitting Part A of Section~\ref{sec:model} -- the model we actually fit, not the target extension of Part B -- to the same neuron. Recall from Section~\ref{sec:model-fitted} that this is
\begin{align}
    & y_{nt} \sim \Pois(\lambda_{nt}),\\
    &\log(\lambda_{nt}) = \alpha + \sum_{i=1}^{13} \sum_{j=1}^{13} \beta_{ijt} X_{nij} + b_{r(n)},
\end{align}
where $n=1,\ldots,3939$, $t=1,\ldots,30$, $\alpha$ is this neuron's fixed intercept (the subscript $k$ dropped as in Section~\ref{sec:model-fitted}, since a single neuron is fixed throughout this section), $b_{r(n)}\sim N(0,1/\tau_b)$ is an i.i.d.\ random effect for the repetition number $r(n)\in\{1,\ldots,13\}$ that trial $n$ belongs to -- the same $13$ repetition labels are shared across all $303$ images, so $r(n)$ indexes which of the $13$ repeated presentations a trial is, not a unique per-trial identifier, and $b_{r(n)}$ absorbs variability associated with repetition number at fixed image and time bin -- and $\beta_{ijt}$ is not a free coefficient per pixel and time bin but is generated by projecting the lower-dimensional SPDE mesh field $\mathbf{w}_t$ onto the pixel grid, $\beta_{ijt}=(\mathbf{A}\mathbf{w}_t)_{ij}$, with $\mathbf{w}_t$ following the AR(1)-in-time, SPDE-in-space structure of Section~\ref{sec:model-fitted}. Estimating this spatio-temporal effect is particularly interesting, as we want to determine which locations of the images spark neural activity more than others, but computation is challenging given the size of the data set and the spatio-temporal variation to be captured -- which is exactly the tractability problem the mesh/projector representation and INLA below are built to solve.

\begin{figure}[H]
    \centering
    \includegraphics[width=0.6\linewidth]{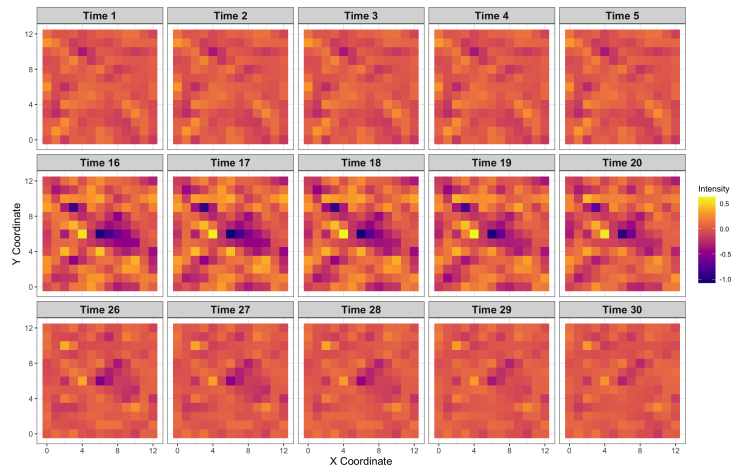}
    \caption{Spatio-temporal varying coefficients capturing the image locations (blue) that spark firing across time in the neuroscience experiment.}
    \label{fig:spatial_brain_images_brain_by_time}
\end{figure}

\paragraph{Why INLA, for this neuron.} Section~\ref{sec:inla-inference} states, in general and with notation, why Part A is fit by INLA rather than MCMC: it is a latent Gaussian model, and INLA's cost scales with the dimension of the four hyperparameters $\bm\psi=(\kappa,\sigma_w,\rho,\tau_b)$ rather than with the dimension of the latent field $\bm\theta$ itself \citep{rue2009approximate,rue2017Bayesian,manna2026scalablespatialpointprocess}. For this neuron specifically: $n=3{,}939$ trials, $T=30$ temporal bins, and a $13\times13$ pixel grid, so the spatial random effect alone projects to on the order of $5{,}070$ correlated pixel-level coefficients (Section~\ref{sec:data}) -- a generic MCMC treatment of a latent field this large is not computationally feasible within a reasonable time frame at this scale, which is exactly the regime Section~\ref{sec:inla-inference}'s argument is built for. We use the SPDE representation of $Q_w$ \citep{bakka2018spatial}, grouped by time with an AR(1) prior; INLA is established methodology, and we do not present its use as the paper's statistical contribution -- the contribution is the structured latent-Gaussian representation of the receptive-field problem that makes an otherwise intractable spatio-temporal fit tractable by this route. Unlike the pixel-level LASSO fit of Section~\ref{sec:empirical-illustration}, the SPDE prior directly encodes that $\beta_{ijt}$ and $\beta_{i'j't}$ should be similar when pixels $(i,j)$ and $(i',j')$ are spatially close; this smoothness discourages, though does not by construction guarantee, the isolated, disconnected selections seen in Figure~\ref{fig:lasso_fit} -- the fit below applies the smoothing (SPDE/AR) component only, without yet the type-level pooling and sparsity mechanism sketched as future work in Section~\ref{sec:future-work}, and Figure~\ref{fig:spatial_brain_images_brain_by_time} should be read accordingly.

\begin{figure}[H]
    \centering
    \includegraphics[width=0.6\linewidth]{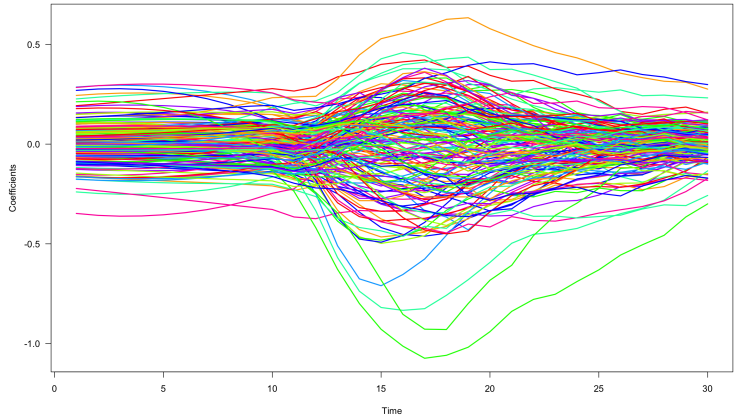}
    \caption{Spatial varying coefficients using SPDE kernel across time in the neuroscience experiment.}
    \label{fig:temporal_brain_images_locations}
\end{figure}

Higher intensity is shown in yellow and lower intensity in blue. The blue region is a point of interest that sparks the neuron more strongly around time interval 15--20, as shown in Figure~\ref{fig:temporal_brain_images_locations}. The spatially varying coefficients appear in Figure~\ref{fig:spatial_brain_images_brain_by_time}. In the rectangle $\{(x,y): 4 \leq x \leq 10,\, 4 \leq y \leq 8\}$, the blue intensity is more prominent in the time window 16--20, where the coefficients fluctuate more than in the other time intervals — the same central spatial region that the empirical illustration of Section~\ref{sec:empirical-illustration} already pointed to, now resolved with an explicit, spatially coherent, time-varying coefficient surface.

This is a demonstration fit to a single neuron, chosen because it lets us walk through the model, the image-level baseline, and the pixel-level LASSO comparison on one concrete case. Section~\ref{sec:population-typing} extends the same per-neuron fit to the full recorded population and uses it to answer the population-level question the hierarchical model in Section~\ref{sec:model} was designed for: whether neurons fall into a small number of distinguishable temporal-response phenotypes.

\subsection{Population-level neuron typing}
\label{sec:population-typing}

The single-neuron fit above answers ``where does this neuron respond?'' The hierarchical model of Section~\ref{sec:model} was motivated by a second question: whether receptive fields cluster into a small number of distinguishable phenotypes, each in principle assignable a type-level structure $\bm\mu_{g,t}$, or whether the population is effectively homogeneous. We investigate this here using the full recorded population of $155$ neurons, as a first, independent-fits step toward the fully joint hierarchical model. We deliberately refer to the resulting clusters as \emph{temporal-response phenotypes} rather than \emph{neuron types} throughout this subsection: they are groups defined by clustering fitted response curves, and treating them as biologically validated cell types would be a stronger claim than the present analysis supports (Section~\ref{sec:limitations}).

For each of the $155$ neurons we repeated the fit of the previous subsection independently, using the same Poisson-response, SPDE-in-space, AR(1)-in-time model, giving each neuron its own spatio-temporal coefficient surface $\widehat{\bm\beta}_{k1},\ldots,\widehat{\bm\beta}_{kT}$. This produces $155$ independently estimated surfaces without yet imposing the shared type-level structure $\bm\mu_{g,t}$ of Section~\ref{sec:model}; typing the population from these surfaces is therefore a two-stage procedure -- independent estimation followed by post hoc clustering -- rather than the fully joint hierarchical fit, and we return to that distinction below.

A clustering analysis of $155$ objects, each a $13\times13\times30 = 5{,}070$-dimensional coefficient surface, first requires choosing a representation. We collapsed each neuron's surface to its space-averaged temporal response profile -- a single curve over the $T=30$ time bins, obtained by averaging the coefficient surface over the $13\times13$ spatial grid at each time point. Concretely, the clustering pipeline has four steps, in the spirit of \citet{jamessugar2003}: (1) expand each neuron's $30$-point temporal profile in a cubic B-spline basis with $8$ basis functions, by least-squares projection, giving an $8$-dimensional coefficient vector per neuron; (2) run principal components analysis on these $155$ basis-coefficient vectors (this is functional PCA under the B-spline basis) and retain the smallest number of components, at least $2$, whose cumulative variance explained reaches $99\%$; (3) fit a Gaussian mixture model (via \texttt{mclust}) to the resulting PCA scores for each candidate number of clusters $k=2,\ldots,10$ and each candidate covariance-model family; (4) select both $k$ and the covariance model by BIC. This selected $k=3$, with cluster sizes $85$, $32$ and $38$ neurons: a well-balanced three-way partition, grouping neurons by the shape of their temporal response, which is exactly the discriminating signal visible when the three clusters' mean curves are plotted together.

\begin{figure}[H]
    \centering
    \includegraphics[width=0.8\linewidth]{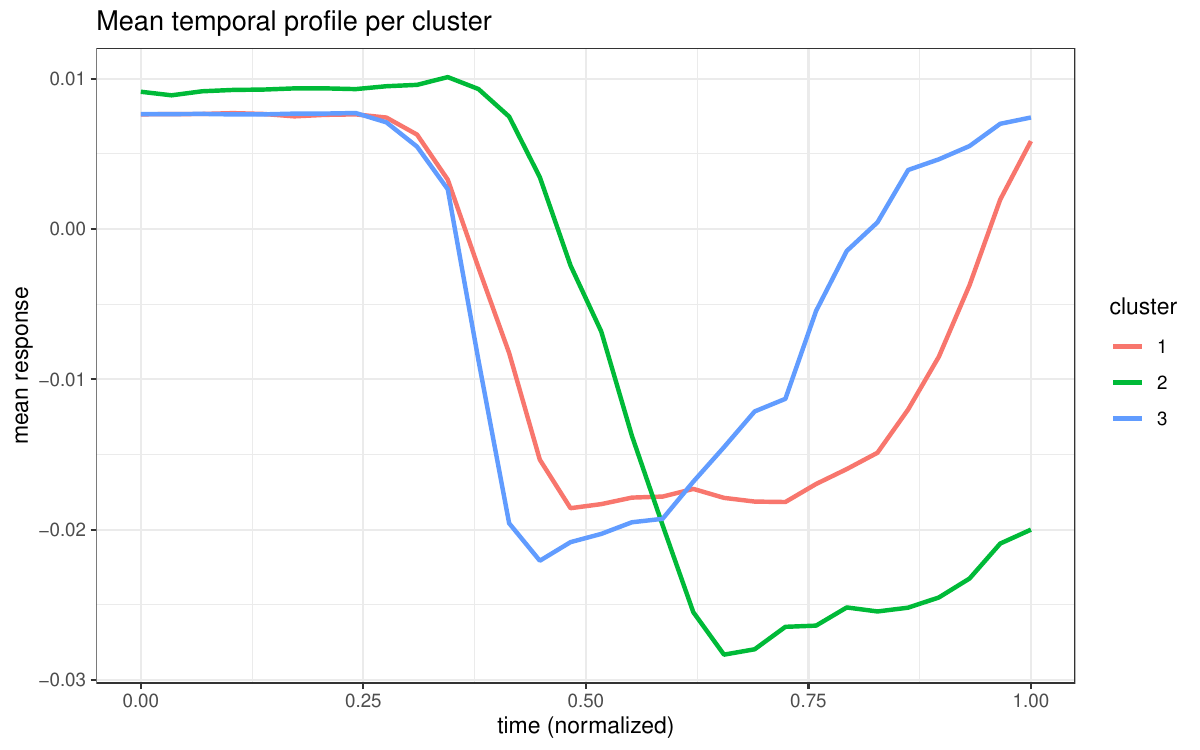}
    \caption{Mean temporal response curve for each of the three temporal-response phenotypes identified by BIC-selected functional clustering of the space-averaged temporal profile ($n=85,32,38$).}
    \label{fig:temporal_cluster_mean_curves}
\end{figure}

\begin{figure}[H]
    \centering
    \includegraphics[width=0.8\linewidth]{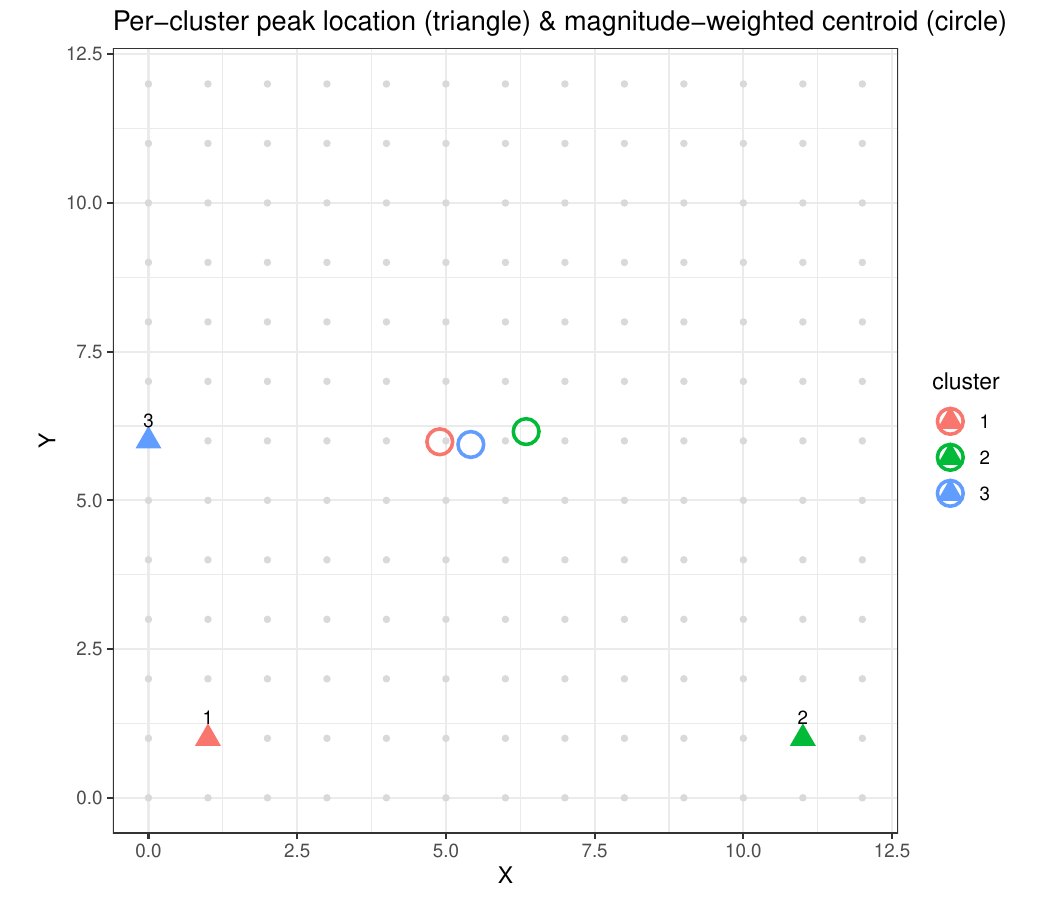}
    \caption{Per-phenotype peak pixel location (triangle) and magnitude-weighted centroid (circle) of the average receptive field, for each of the three temporal-response phenotypes.}
    \label{fig:cluster_location_map_temporal}
\end{figure}

\paragraph{Characterizing the three phenotypes.} For the primary, temporal-profile-based partition (sizes $85$, $32$, $38$), we examined the mean temporal response curve within each cluster (Figure~\ref{fig:temporal_cluster_mean_curves}) and the spatial location of each cluster's peak and magnitude-weighted-centroid response (Figure~\ref{fig:cluster_location_map_temporal}), computed as follows. For each of the three clusters, we averaged, pixel by pixel and time bin by time bin, the fitted $13\times13\times30$ coefficient surfaces $\widehat{\bm\beta}_{k1},\ldots,\widehat{\bm\beta}_{kT}$ of its member neurons, giving one cluster-average spatio-temporal field -- this is a full $13\times13\times30$ object, not the space-averaged temporal profile used for clustering above. Within that cluster-average field we then located the single pixel-by-time-bin cell of largest absolute magnitude; its spatial coordinates are the ``peak'' (triangle, Figure~\ref{fig:cluster_location_map_temporal}) and its time bin is the ``peak time.'' The ``magnitude-weighted centroid'' (circle, Figure~\ref{fig:cluster_location_map_temporal}) is computed only at that one peak time bin: it is the spatial average of all $169$ pixel coordinates, each weighted by the absolute value of the cluster-average field at that pixel at the peak time bin -- not a centroid integrated across time, and not restricted to a thresholded active set. Absolute value is used throughout this computation because the fitted coefficient can be positive (excitatory) or negative (inhibitory), and both are equally a departure from baseline for the purpose of locating where a cluster's response is strongest.

All three phenotypes share the same broad temporal shape -- a positive response over roughly the first third of the trial, a decline into negative territory around the middle third, and a partial late recovery -- but differ systematically in the timing and depth of this decline (Figure~\ref{fig:temporal_cluster_mean_curves}). Phenotype 2 ($n=32$) declines earliest and most sharply, reaching the most negative mean response of the three by roughly two-thirds of the way through the trial, and recovers only partially by the end. Phenotype 3 ($n=38$) declines slightly earlier than Phenotype 1 but recovers to the highest final value of the three. Phenotype 1 ($n=85$), the largest and most heterogeneous group, declines latest and most gradually and recovers to an intermediate final level. We considered showing the cluster-average spatio-temporal field at a sequence of time bins as a complementary visualization, but dropped it: at this resolution the three phenotypes' spatial fields were visually indistinguishable from one another by the middle and late time bins, so the figure conveyed no information beyond the mean-curve differences already shown in Figure~\ref{fig:temporal_cluster_mean_curves}, and we do not report a figure we would not ourselves find informative.

Spatially, the magnitude-weighted centroids of the three phenotypes' average receptive fields (circles, Figure~\ref{fig:cluster_location_map_temporal}) fall close together near the center of the pixel grid, indicating that the three phenotypes are not primarily distinguished by the gross location of their receptive field. Each phenotype's single peak-intensity pixel (triangles, Figure~\ref{fig:cluster_location_map_temporal}) falls near the grid's periphery, but at three different peripheral locations; because this is a single-pixel summary rather than a weighted regional one, we read it as suggestive rather than conclusive, and do not treat it as evidence of a strong spatial dissociation between phenotypes beyond the temporal-dynamics differences documented above. Taken together, the population typing recovered here is best described as three temporal-response dynamics -- an early-and-sharp decline, a late-and-gradual decline, and an intermediate pattern -- shared by neurons whose average receptive fields occupy broadly the same central visual region. We have not tested whether these phenotypes correspond to independently established retinal ganglion cell classes (e.g., ON/OFF or fast/slow types from the physiology literature); doing so would require comparing against such labels directly and is left to future work.

\paragraph{Relationship to the joint hierarchical model.} The typing above uses independent per-neuron fits followed by clustering, rather than the joint hierarchical model of Section~\ref{sec:model} fit directly to all $155$ neurons with a shared $\bm\mu_{g,t}$ per phenotype. This two-stage procedure is a practical first step rather than a substitute for the joint fit: it gives a data-driven value of $G$ (informed by the $85/32/38$ partition found here) to use when fitting the joint model, but it does not pool information across neurons within a phenotype, and so does not itself demonstrate what that pooling would add. Fitting the joint model with $G=3$ and quantifying that gain is future work, detailed in Section~\ref{sec:future-work}.

\section{Interpreting the fitted coefficients}
\label{sec:interpretation}

The sign of the coefficient surface fit in Part A already supports a direct scientific reading: for neuron $k$, $\beta_{kst}>0$ indicates an excitatory association between pixel $s$ and firing at time $t$, and $\beta_{kst}<0$ an inhibitory one, which is exactly the yellow/blue reading used in the real-data figures of Section~\ref{app:real-data}. INLA's marginal posteriors also give, for free, the neuron-level probability $P(\beta_{kst}>0\mid Y)$ at every pixel and time bin -- a direct posterior measure of evidence for excitation or inhibition at that location, available today from Part A alone, with no extension required. The type-level analogues of these quantities -- a population-level field $\bm\mu_{g,t}$ per phenotype, and a posterior probability of type-level membership $P(\mu_{gst}>0\mid Y)$ -- are properties of the not-yet-fit extension of Section~\ref{sec:future-work}, and are deferred there rather than stated here as if already available.

\section{Simulation Study}
\label{sec:simulation}

Section~\ref{sec:interpretation} and Section~\ref{app:real-data} interpret the fitted coefficients on real neurons, but a real neuron comes with no known true receptive field: nobody knows in advance which pixels genuinely drive a real neuron's firing, so there is no ground truth to check any method's estimate against. On real data we can therefore only ask how well a method \emph{predicts held-out spikes} (the log-score) and what its fitted coefficient surface \emph{looks like} (how many pixels it calls active, whether they form one connected region or many); we cannot ask whether it found the \emph{right} pixels, because ``right'' is not known. A good log-score means a method's predicted firing rates track the held-out spike counts well on average -- it rewards getting the overall pattern of activity right, but a method can achieve a good log-score while still misattributing exactly which pixels matter, as long as its overall predictions are accurate; it is silent on support recovery by construction, not by an oversight in how we reported it. This is precisely the gap a simulation study closes: by generating synthetic neurons from a receptive field \emph{we choose}, the true active pixels are known by construction, and criteria that are impossible to compute on real data -- true-positive rate (TPR), false-positive rate (FPR), F1, and coefficient estimation error $\|\widehat{\bm\beta}-\bm\beta\|_2$ -- become computable. On real data, only the held-out log-score and the qualitative structure of the fitted coefficient surface are available (Section~\ref{sec:real-data}); this section reports the simulation comparison instead, with all of the above computable because the truth is known -- the paper's primary quantitative evidence for how the proposed model compares against simpler non-hierarchical alternatives.

\subsection{Design}
\label{sec:sim-design}

The design this study is built toward is broader than what has been executed. In full, true receptive fields would be drawn from five spatial configurations -- compact contiguous regions, multiple disconnected regions, mixed-sign regions, spatially smooth (uncut) regions, and temporally changing regions -- crossed with the full range $G\in\{1,2,3\}$ and a continuum of deviation magnitudes $\tau_\delta$ (rather than two fixed levels), with the neuron-specific deviation drawn as a literal Gaussian Markov random field, $\bm\delta_{kt}\sim N(\mathbf{0},\tau_\delta^2Q_\delta^{-1})$, matching the target decomposition of Section~\ref{sec:future-work} exactly (Section~\ref{sec:sim-datagen} explains where the executed version below departs from this). What follows is the reduced slice of that design actually implemented and executed, in \texttt{simulation\_study.R}; it is smaller for a stated, practical reason (SPDE/AR(1) fits are slow enough per neuron that the full factorial is not yet feasible), not a silent one. We vary three axes factorially:
\begin{itemize}[leftmargin=2em]
\item \textbf{spatial configuration} $\in\{\text{compact},\ \text{disconnected}\}$ -- the two of the five above most directly relevant to this paper's central question, whether a method returns one coherent region (compact) or correctly keeps two separate regions apart rather than merging or fragmenting them (disconnected);
\item \textbf{true number of neuron types} $G\in\{1,3\}$ -- $G=1$, a null case where the population is genuinely homogeneous; $G=3$, matching the value found by the real-data clustering of Section~\ref{sec:population-typing} ($G=2$ is part of the full design above but not simulated here);
\item \textbf{deviation magnitude} $\tau_\delta\in\{0.4\ (\text{low}),\ 1.6\ (\text{high})\}$ -- how far an individual neuron's true field is allowed to drift from its type's field.
\end{itemize}
This gives $2\times2\times2=8$ combinations. For each, $K=12$ synthetic neurons are generated on the same $13\times13$ ($P=169$) pixel grid as the real data, over $T=6$ time bins (reduced from the real data's $T=30$ for computational tractability -- SPDE/AR(1) fits scale with $T$, and $6$ bins is enough to exercise the AR(1) linkage without the full real-data run time).

\subsection{Ground-truth receptive fields: exactly which pixels are active, and why}
\label{sec:sim-groundtruth}

Every true field below is built from one primitive,
\begin{equation}
g(x,y;x_0,y_0,\sigma,a) = \begin{cases} a\,\exp\!\left(-\dfrac{(x-x_0)^2+(y-y_0)^2}{2\sigma^2}\right) & (x-x_0)^2+(y-y_0)^2 \le (2.5\sigma)^2, \\[4pt] 0 & \text{otherwise,} \end{cases}
\end{equation}
a Gaussian bump centered at $(x_0,y_0)$ with amplitude $a$ and spread $\sigma$, \textbf{hard-zeroed outside a radius of $2.5\sigma$}. This cutoff is the essential design choice, and the reason for it is worth stating precisely: a plain, uncut Gaussian is nonzero at \emph{every} pixel, however far from the center, so ``the true active region'' would only ever be a post-hoc thresholding convention applied to a field with no true zeros -- exactly the flaw a simulation study must avoid if it is going to test support recovery honestly. Cutting at $2.5\sigma$ instead gives pixels outside the radius \emph{exactly} $0$, a real ground truth, while keeping the discontinuity small: at $d=2.5\sigma$ a plain Gaussian is already down to $\exp(-2.5^2/2)\approx4.4\%$ of its peak, so the cutoff removes only what a reasonable threshold would have called negligible anyway. Formally, the true-active mask for a field $\bm\beta$ built this way is $\{s : |\beta_s| > 0\}$ exactly -- not a $5\%$-of-maximum threshold, which is instead reserved for scoring every method's \emph{estimated} coefficients (Section~\ref{sec:sim-evaluation}), none of which ever reach exact zero by construction of GLM, LASSO, elastic net, or the SPDE smoother alike.

Three fixed type centers are used (reused across both configurations, so ``type'' is a genuine, fixed spatial identity): type 1 near $(4,4)$/$(10,10)$, type 2 near $(10,4)$/$(4,10)$, type 3 near $(7,10)$/$(7,4)$ (grid coordinates on the $13\times13$ pixel grid). Concretely:
\begin{itemize}[leftmargin=2em]
\item \textbf{compact}: one bump per type, $\sigma=1.6$, amplitude $1.6$ -- a single contiguous active region, roughly $13$-$14$ pixels across at half-maximum, surrounded by exact zeros.
\item \textbf{disconnected}: two bumps per type at the type's two listed centers, $\sigma=1.1$, amplitudes $+1.6$ and $-1.6$ -- two separate active regions of opposite sign, with exact zero everywhere between and around them.
\end{itemize}
Every bump is modulated in time by a fixed envelope $e(t) = \sin\!\big(\pi(t-0.5)/T\big)$, $t=1,\ldots,T$, which rises from near $0$ to $1$ and back down -- so the \emph{set} of active pixels does not change over time (a pixel that is ever active is active at every $t$, just more or less strongly), only the magnitude does. For a representative compact-configuration type field, this construction gives $47$ pixels genuinely active (nonzero) and $122$ exactly zero out of $169$; for disconnected, $42$ active and $127$ exactly zero -- the precise split depends on the type and configuration, but every field has a real, checkable answer, which is the point.

% \begin{figure}[H]
% \centering
% \includegraphics[width=0.85\linewidth]{figures/simulation_true_fields.pdf}
% \caption{Target (ground-truth) receptive fields used in the simulation study: compact vs.\ disconnected spatial configurations (rows) crossed with the three fixed neuron types (columns), shown as the time-averaged spatial shape (this section); amplitude at each time bin is additionally scaled by the temporal envelope $e(t)$ shown beneath the fields. White pixels are exact zeros, not small values -- the hard cutoff at $2.5\sigma$ described above.}
% \label{fig:sim_true_fields}
% \end{figure}

\subsection{Neurons and data generation}
\label{sec:sim-datagen}

Each of the $K=12$ neurons per combination is assigned a type $g(k)\in\{1,\ldots,G\}$ (cycled evenly across neurons), and its true field is
\begin{equation}
\bm\beta_{kt} = \bm\mu_{g(k),t} + \bm\delta_{kt},
\end{equation}
matching the decomposition of Section~\ref{sec:future-work}. \emph{We flag one honest simplification here.} Section~\ref{sec:future-work}'s target design draws $\bm\delta_{kt}$ from a Gaussian Markov random field, $\bm\delta_{kt}\sim N(\mathbf{0},\tau_\delta^2 Q_\delta^{-1})$ -- fine-grained, spatially correlated noise. What is actually implemented is simpler: $\bm\delta_{k\cdot}$ is one more hard-cutoff bump (Section~\ref{sec:sim-groundtruth}), at a random center drawn uniformly on the grid interior, random sign, amplitude $\tau_\delta$, constant across $t$. This is a reasonable proxy for ``individual deviation from the type'' -- it is still smooth, still genuinely localized, and its magnitude is still controlled by $\tau_\delta$ -- but it is not literally a GMRF draw, and we say so plainly rather than let the two be confused: implementing the literal GMRF version is a small, well-defined piece of remaining work, not yet done.

Each combination draws $n=220$ synthetic ``images,'' i.i.d.\ $X_{np}\sim N(0,0.35^2)$ per pixel ($220>169=P$ so the pooled design matrix is full rank, mirroring why the real data's $303>169$ images avoid the same issue), shared across all $12$ neurons in that combination. For neuron $k$, image $i$, time bin $t$,
\begin{equation}
\eta_{it} = \alpha_k + \mathbf{x}_i^\top\bm\beta_{kt}, \qquad \lambda_{it} = \exp(\eta_{it}), \qquad Y_{itr}\mid\lambda_{it} \iid \Pois(\lambda_{it}), \quad r=1,\ldots,4,
\end{equation}
with $\alpha_k = \log(2.5) + \varepsilon_k$, $\varepsilon_k\sim N(0,0.1^2)$, and $4$ repeated presentations per image, matching the real data's repeated-presentation design at smaller scale. Of the $4$ repetitions, $\mathbf{1}$ is held out for evaluation and the other $\mathbf{3}$ used for training -- a $25\%$ held-out fraction, in the same spirit as the real data's repeated-presentation design ($13$ presentations per image; Section~\ref{sec:data}). (We note this correction explicitly: an earlier draft of this section, and the code comment that accompanied it, described this the other way around -- $3$ of $4$ held out -- which was backwards; the actual implementation, and every number reported below, holds out $1$ of $4$.) One further implementation detail, checked directly rather than assumed: the implementation clips $\eta_{it}$ to $[-6,5]$ before exponentiating, as a numerical-overflow guard. Replaying the generator with this clip removed shows it binds rarely at $\tau_\delta=0.4$ (under $0.2\%$ of generated observations in every combination) but more often at $\tau_\delta=1.6$ (up to $3.7\%$, in the disconnected/$G=1$ combination); across all $8$ combinations combined, $1.57\%$ of the $126{,}720$ generated linear predictors were clipped. This affects only the upper tail of $\lambda_{it}$ in the high-$\tau_\delta$ combinations and is small enough not to change any qualitative comparison above, but it means $\eta_{it}=\alpha_k+\mathbf{x}_i^\top\bm\beta_{kt}$ is not exactly what generated every observation at $\tau_\delta=1.6$; we report the discrepancy rather than omit it.

\subsection{Methods and evaluation}
\label{sec:sim-evaluation}

The full comparison this design targets is six methods: (1) unregularized Poisson GLM, (2) Poisson-LASSO, (3) Poisson elastic net, (4) spatial-only SPDE, (5) spatio-temporal SPDE/AR(1) (Part A of Section~\ref{sec:model-fitted}), and (6) the proposed hierarchical structured-sparsity model (Section~\ref{sec:future-work}); comparing (3) against (6) asks whether accounting for predictor correlation alone is enough, and (5) against (6) isolates the benefit of hierarchical type-sharing specifically. Methods (1)--(3) are fit for all $12$ neurons per combination, coefficients pooled across time bins (as in Section~\ref{app:real-data}'s real-neuron fit). Methods (4)--(5) (fit per neuron with no type-sharing) are fit via INLA for the first $4$ of the $12$ neurons per combination only, since SPDE/AR(1) fits are substantially slower; all $4\times8=32$ fits, for both methods, converged (zero failures). Method (6) is not implemented and not run here -- it has no native INLA specification, and we do not write code for an unvalidated method in advance of a real design and fitting plan. Every method's coefficient surface is compared, at the same $5\%$-of-maximum active-set threshold used throughout, against the time-averaged true coefficient $\bar\beta_{ks}=\frac{1}{T}\sum_t\beta_{kst}$ (since methods (1)--(3) estimate one time-pooled coefficient, not a time-varying one), via: held-out Poisson log-score; $\|\widehat{\bm\beta}-\bm\beta\|_2$; TPR, FPR, and F1 against the true-active mask of Section~\ref{sec:sim-groundtruth}; and the number of $4$-connected components in the estimated active set. Three further evaluation dimensions belong to the full design above but are not part of this executed slice: a formal comparison of estimated versus true connected-component counts, rather than the raw estimated count already computed; temporal stability, whether an estimated field evolves smoothly when the true field does; and uncertainty calibration, the empirical coverage of INLA's posterior credible intervals. These remain future work alongside the joint hierarchical fit (Section~\ref{sec:future-work}).

\subsection{Results}
\label{sec:sim-results}

\emph{A note on how these numbers were obtained.} The tables below come from re-running \texttt{simulation\_study.R} end to end, with INLA installed, under the corrected hard-cutoff ground-truth design of Section~\ref{sec:sim-groundtruth}; no other part of the design, data-generation, or evaluation code (Sections~\ref{sec:sim-design}--\ref{sec:sim-evaluation}) changed. An earlier run, made before that correction (a plain, uncut Gaussian with no true zeros -- exactly the flaw Section~\ref{sec:sim-groundtruth} explains), produced a first version of these tables; that version is superseded and not reported here. We compared the two runs directly, method by method and combination by combination: log-score and coefficient error are essentially unchanged (mean absolute differences of $0.02$ and $0.03$ respectively, over the $40$ method$\times$combination cells in both tables), and the qualitative ranking among methods on every criterion is the same in both runs. TPR is the one metric materially affected, since it depends directly on what ``truly active'' means: it drops for every method under the stricter, exact-zero ground truth (mean absolute difference $0.06$, up to $0.15$ in the worst case), because the earlier, uncut ground truth counted many near-zero, never-truly-inactive pixels as true positives that the corrected definition now correctly excludes. FPR is affected only modestly (mean absolute difference $0.02$).

\begin{table}[H]
\centering
\footnotesize
\caption{Simulation results, compact spatial configuration, computed under the corrected hard-cutoff ground truth of Section~\ref{sec:sim-groundtruth}. GLM, LASSO, and Elastic net rows average over $12$ neurons per combination; Spatial SPDE and Spatio-temporal SPDE/AR(1) rows average over the first $4$ (Section~\ref{sec:sim-evaluation}). TPR/FPR/coefficient error are computed against the true coefficient surface, at the same active-set threshold used throughout the paper -- see the note preceding this table for a comparison against an earlier, superseded run.}
\begin{tabular}{llrrrrr}
\toprule
$G$ & $\tau_\delta$ & Method & Log-score & TPR & FPR & Coef.\ error \\
\midrule
1 & high & GLM & $-2.966$ & $0.88$ & $0.57$ & $2.15$ \\
1 & high & LASSO & $-3.024$ & $0.73$ & $0.25$ & $1.79$ \\
1 & high & Elastic net & $-3.041$ & $0.73$ & $0.26$ & $1.93$ \\
1 & high & Spatial SPDE & $-2.595$ & $0.92$ & $0.27$ & $1.04$ \\
1 & high & Spatio-temporal SPDE/AR(1) & $-1.845$ & $0.93$ & $0.23$ & $0.95$ \\
1 & low & GLM & $-2.371$ & $0.83$ & $0.52$ & $1.08$ \\
1 & low & LASSO & $-2.409$ & $0.56$ & $0.12$ & $1.01$ \\
1 & low & Elastic net & $-2.403$ & $0.60$ & $0.15$ & $1.01$ \\
1 & low & Spatial SPDE & $-2.324$ & $0.82$ & $0.12$ & $0.50$ \\
1 & low & Spatio-temporal SPDE/AR(1) & $-1.803$ & $0.82$ & $0.07$ & $0.39$ \\
3 & high & GLM & $-2.822$ & $0.88$ & $0.61$ & $2.10$ \\
3 & high & LASSO & $-2.886$ & $0.73$ & $0.23$ & $1.98$ \\
3 & high & Elastic net & $-2.887$ & $0.74$ & $0.25$ & $2.02$ \\
3 & high & Spatial SPDE & $-3.001$ & $0.89$ & $0.56$ & $1.78$ \\
3 & high & Spatio-temporal SPDE/AR(1) & $-2.035$ & $0.89$ & $0.49$ & $1.65$ \\
3 & low & GLM & $-2.570$ & $0.85$ & $0.57$ & $1.16$ \\
3 & low & LASSO & $-2.620$ & $0.56$ & $0.10$ & $1.09$ \\
3 & low & Elastic net & $-2.612$ & $0.59$ & $0.12$ & $1.06$ \\
3 & low & Spatial SPDE & $-2.629$ & $0.81$ & $0.08$ & $0.52$ \\
3 & low & Spatio-temporal SPDE/AR(1) & $-1.912$ & $0.83$ & $0.04$ & $0.41$ \\
\bottomrule
\end{tabular}
\label{tab:sim-compact}
\end{table}

\begin{table}[H]
\centering
\footnotesize
\caption{Simulation results, disconnected spatial configuration, same design, columns, and ground-truth definition as Table~\ref{tab:sim-compact}.}
\begin{tabular}{llrrrrr}
\toprule
$G$ & $\tau_\delta$ & Method & Log-score & TPR & FPR & Coef.\ error \\
\midrule
1 & high & GLM & $-2.858$ & $0.87$ & $0.58$ & $1.98$ \\
1 & high & LASSO & $-2.929$ & $0.73$ & $0.16$ & $1.83$ \\
1 & high & Elastic net & $-2.933$ & $0.73$ & $0.19$ & $1.90$ \\
1 & high & Spatial SPDE & $-2.898$ & $0.91$ & $0.40$ & $1.41$ \\
1 & high & Spatio-temporal SPDE/AR(1) & $-2.013$ & $0.91$ & $0.37$ & $1.32$ \\
1 & low & GLM & $-2.319$ & $0.85$ & $0.48$ & $1.03$ \\
1 & low & LASSO & $-2.366$ & $0.61$ & $0.07$ & $0.92$ \\
1 & low & Elastic net & $-2.360$ & $0.63$ & $0.09$ & $0.92$ \\
1 & low & Spatial SPDE & $-2.361$ & $0.82$ & $0.26$ & $0.88$ \\
1 & low & Spatio-temporal SPDE/AR(1) & $-1.856$ & $0.81$ & $0.24$ & $0.84$ \\
3 & high & GLM & $-2.774$ & $0.89$ & $0.58$ & $1.79$ \\
3 & high & LASSO & $-2.846$ & $0.74$ & $0.19$ & $1.73$ \\
3 & high & Elastic net & $-2.850$ & $0.75$ & $0.22$ & $1.80$ \\
3 & high & Spatial SPDE & $-2.795$ & $0.92$ & $0.50$ & $1.54$ \\
3 & high & Spatio-temporal SPDE/AR(1) & $-2.001$ & $0.92$ & $0.42$ & $1.46$ \\
3 & low & GLM & $-2.744$ & $0.82$ & $0.47$ & $1.03$ \\
3 & low & LASSO & $-2.813$ & $0.55$ & $0.10$ & $1.02$ \\
3 & low & Elastic net & $-2.826$ & $0.55$ & $0.12$ & $1.08$ \\
3 & low & Spatial SPDE & $-2.813$ & $0.85$ & $0.36$ & $1.01$ \\
3 & low & Spatio-temporal SPDE/AR(1) & $-1.984$ & $0.84$ & $0.33$ & $0.89$ \\
\bottomrule
\end{tabular}
\label{tab:sim-disconnected}
\end{table}

\begin{figure}[p]
\centering
\includegraphics[width=\linewidth]{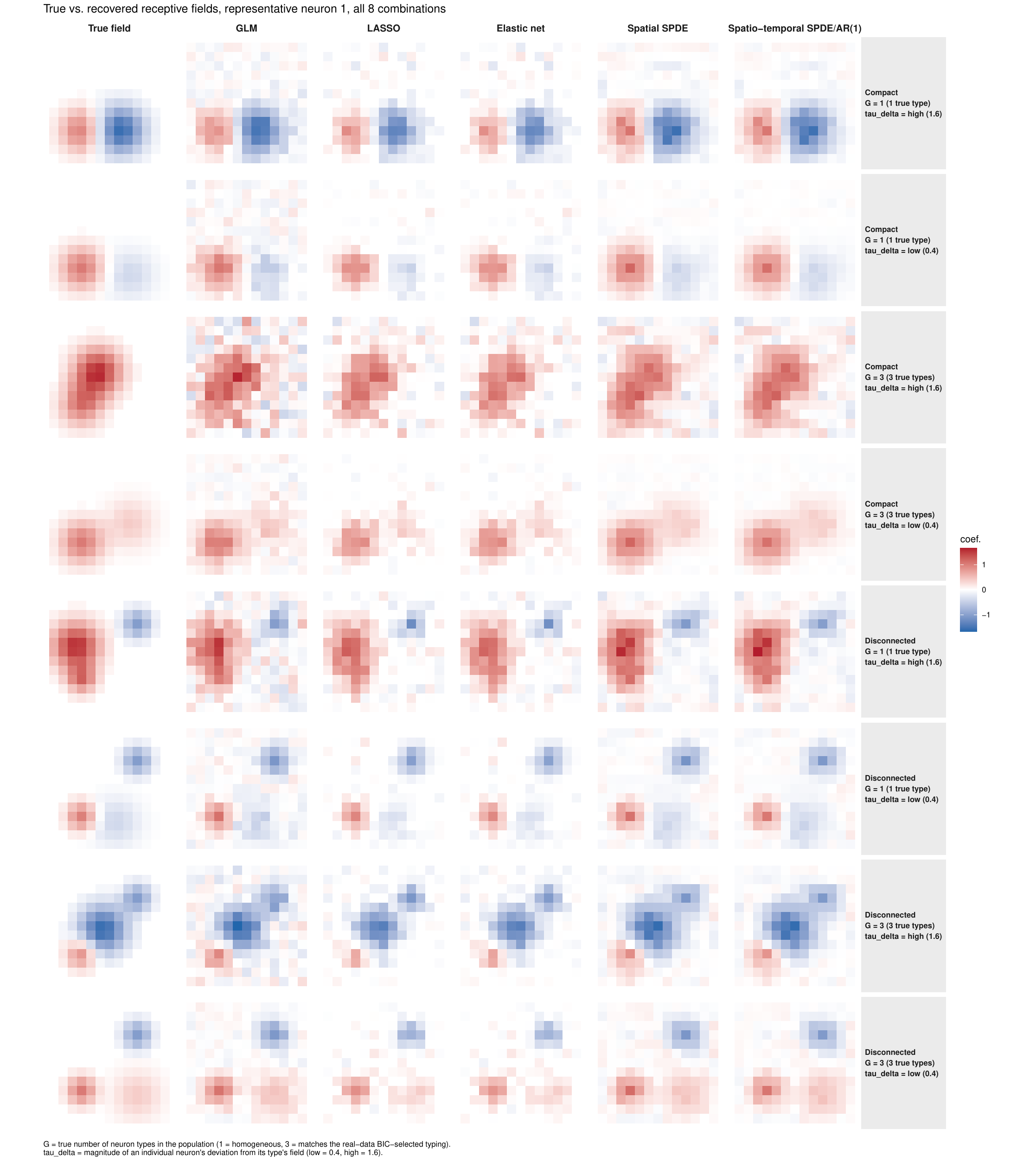}
\caption{True vs.\ recovered receptive fields for a representative neuron (neuron $1$ of $12$, selected by a fixed index specified in advance of the analysis) in each of the $8$ simulation combinations (rows), across the true field and all five fitted comparison methods (columns). Rows are labeled by spatial configuration, true number of neuron types $G$, and deviation magnitude $\tau_\delta$ (both its low/high name and numeric value); color encodes the fitted coefficient at each pixel, centered at zero (white), with blue/red indicating negative/positive values. The LASSO and elastic-net panels visibly reach exact zero over large contiguous areas, matching their design; the SPDE/AR(1) panels are visibly smoother and better localized than the unregularized GLM but, consistent with Tables~\ref{tab:sim-compact}--\ref{tab:sim-disconnected}, retain small nonzero values outside the true active region rather than hard zeros.}
\label{fig:sim_compare}
\end{figure}

\paragraph{Interpretation.} Averaging across all $8$ combinations (both tables), spatio-temporal SPDE/AR(1) wins on log-score in $8$ of $8$ combinations (mean $-1.93$, versus $-2.68$ for GLM and $-2.74$ for LASSO) and on coefficient error in $8$ of $8$ (mean $0.99$, versus $1.42$ for LASSO and $1.54$ for GLM). This is a real, consistent advantage on two of four criteria, not a marginal one. On the other two criteria the picture is more mixed than a first, uncorrected pass at these numbers suggested. TPR is highest, on average, for spatio-temporal SPDE/AR(1) (mean $0.869$, against $0.859$ for GLM and $0.650$ for LASSO), but under the corrected, exact-zero ground truth it no longer wins the \emph{individual-combination} count: GLM has the highest TPR in $3$ of $8$ combinations, spatial SPDE in $3$ of $8$, and spatio-temporal SPDE/AR(1) in only $2$ of $8$ -- the three are close enough, combination by combination, that no single method dominates. This is the metric most affected by the ground-truth correction (Section~\ref{sec:sim-results}'s note above): the earlier, uncut ground truth had systematically inflated every method's apparent TPR by counting near-zero pixels as true positives, and correcting it changes not just the numbers but which method looks best on this one criterion. On false-positive control, LASSO has the lowest (best) FPR in $5$ of $8$ combinations (mean $0.152$) and spatio-temporal SPDE/AR(1) in the remaining $3$ (mean $0.276$ overall) -- LASSO's advantage here is real but no longer as one-sided as before. None of this changes the paper's central, and most robust, finding: LASSO is designed to force coefficients to exact zero, while the SPDE/AR(1) fit, as actually implemented here, only smooths and never reaches exact zero, so near a true region's edge it tends to leave small-but-above-threshold values that count as false positives once thresholded. This is precisely the gap the structured-sparsity mechanism of Section~\ref{sec:future-work} is designed to close on top of the smoothing tested here; the simulation's clearest, most consistent evidence for that gap remains the predictive and coefficient-accuracy comparison (log-score and coefficient error, both won in $8$ of $8$ combinations), not the more evenly split support-recovery comparison. Spatial SPDE alone (method (4), no temporal linkage) beats the GLM on coefficient error throughout but only beats it on log-score in $2$ of $8$ combinations; the temporal AR(1) linkage is what turns spatial smoothing into a consistently better predictor, not spatial smoothing on its own. A visual comparison of the true and recovered fields across all $8$ combinations is shown in Figure~\ref{fig:sim_compare}; a further check of whether SPDE smoothing helps at all relative to a non-spatial baseline is the GLM-versus-SPDE comparison itself, and the answer throughout is a qualified yes: both SPDE variants beat the GLM on coefficient error in every combination, but only the spatio-temporal variant, with the AR(1) linkage across the $T=6$ time bins, translates that into a consistently better held-out predictive score.

\section{Discussion}

The central premise of this work is that receptive-field estimation should be treated as a structured statistical problem rather than simply as a high-dimensional variable-selection problem.

LASSO provides an attractive baseline because it produces sparse estimates and is computationally accessible. However, its treatment of predictors does not directly encode the fact that image pixels have spatial relationships. This distinction is important when the scientific objective is to identify coherent receptive-field regions. Previous statistical work has documented limitations of LASSO under dependence among covariates, while the elastic net provides one established approach for inducing grouping among correlated predictors.

The proposed framework goes beyond grouping alone. It simultaneously considers spatial structure, temporal evolution, neuron-specific heterogeneity, and sharing across neuron types.

The model also creates an opportunity to distinguish two different scientific questions:
\begin{equation}
\text{``Where does this neuron respond?''} \qquad \text{and} \qquad \text{``How does this response differ across neuron types?''}
\end{equation}
The first question concerns individual receptive fields, whereas the second concerns population-level organization. A hierarchical model allows both questions to be addressed within a unified framework: rather than choosing between treating all neurons as identical or treating them as completely independent, the decomposition $\beta_{kst} = \mu_{g(k),st} + \delta_{kst}$ (Section~\ref{sec:future-work}, not yet fit) lets information be shared within a neuron type while preserving neuron-specific heterogeneity. If, say, one neuron within a functional type has relatively noisy observations, its receptive-field estimate can borrow strength from the other neurons of the same type, while a neuron belonging to a different type remains free to have a different population-level receptive field.

\paragraph{Methodological note.} Clustering the raw $5{,}070$-dimensional per-neuron coefficient surfaces directly, before settling on the space-averaged temporal-profile representation used in Section~\ref{sec:population-typing}, did not work well: it gave a degenerate, outlier-dominated split rather than a usable population typing.

\paragraph{Limitations and scope.}
\label{sec:limitations}
This section states plainly what the present analysis does and does not establish.

\begin{enumerate}[label=(\roman*), leftmargin=2em]
\item \emph{The joint hierarchical model is sketched as a direction, not specified in full or fit.} Section~\ref{sec:future-work} states the decomposition $\beta_{kst}=\mu_{g(k),st}+\delta_{kst}$ that a joint fit would target, but does not work out a complete prior specification for it; the empirical results of Section~\ref{app:real-data} come entirely from the per-neuron SPDE/AR(1) smoothing component fit independently, followed by post hoc clustering, not from any joint fit across neurons. This is the paper's central limitation. Two consequences follow directly: spatial smoothing alone discourages but does not formally guarantee connected sparse support, which a type-level sparsity mechanism would need to address -- Section~\ref{sec:sim-results}'s simulation already shows this gap is real, not hypothetical; and the benefit of pooling information across neurons within a type is correspondingly unmeasured, since only independent fits have been run. Working out and fitting the joint model with $G=3$ (informed by Section~\ref{sec:population-typing}), and comparing the resulting $\widehat{\bm\mu}_{g,t}$ and pooled $\widehat{\bm\delta}_{kt}$ against the independent per-neuron fits used here, is the single highest-priority next step, sketched briefly in Section~\ref{sec:future-work}.
\item \emph{The three temporal-response phenotypes are an empirical partition, not a validated biological typing.} They are referred to as phenotypes rather than neuron types throughout Section~\ref{sec:population-typing} because they have not been compared against an independently established biological classification. The temporal-profile representation used to obtain them is motivated substantively (space-averaging removes neuron-specific spatial idiosyncrasy while retaining response dynamics), but its robustness to the choice of spatial summary, clustering algorithm, model-selection criterion, or SPDE hyperparameters has not been tested.
\item \emph{A quantitative, held-out comparison of the proposed model against simpler non-hierarchical alternatives on real data (Poisson GLM, LASSO, elastic net) is not reported in this paper.} On a real neuron the true active set is never known, so a real-data predictive comparison alone cannot be read as evidence about support recovery, and we do not include one here. The simulation study of Section~\ref{sec:simulation}, where the true receptive field is known, is the paper's primary quantitative evidence that the proposed spatio-temporal smoothing recovers support and reduces coefficient error relative to these alternatives, though not uniformly: LASSO controls false positives better (Section~\ref{sec:sim-results}). Evaluating real-data predictive accuracy against these alternatives, together with the sixth comparison -- against the proposed model's own hierarchical, type-sharing extension, once fit -- remains future work (Section~\ref{sec:future-work}).
\item \emph{The result has not been replicated.} It comes from a single experimental preparation (one set of salamander retinal ganglion cell recordings under one stimulus set); an independent preparation, or a different sensory system, would be needed before treating the three-phenotype structure as a general finding rather than a property of this dataset.
\end{enumerate}

Further extensions are possible beyond closing the gaps above: a negative-binomial likelihood to accommodate overdispersion, neural history terms for temporal dependence in spike generation beyond the receptive field itself, a more flexible nonparametric spatial prior for continuous stimulus geometries, and theoretical analysis of posterior concentration and support recovery under spatially dependent predictors. The same spatio-temporally structured estimation principle is not specific to retinal ganglion cells: population receptive-field mapping in human visual cortex \citep{dumoulinwandell2008} faces an analogous problem of estimating a spatially coherent field per voxel from noisy responses, and whether the present approach transfers to that setting is a natural question for a human-experiment follow-up, should one become available.

\section{Future Work}
\label{sec:future-work}

Every result in this paper comes from fitting the per-neuron model of Part A (Section~\ref{sec:model-fitted}) to the $155$ recorded neurons independently, combining information across them only afterward, through the post hoc clustering of Section~\ref{sec:population-typing}. The natural next step is to pool information across neurons of the same functional type directly inside the model, rather than only after each neuron's field has already been estimated on its own. Concretely, this would decompose each neuron's receptive field into a shared type-level component and a neuron-specific deviation,
\begin{equation}
\beta_{kst} = \mu_{g(k),st} + \delta_{kst}, \qquad \bm\delta_{kt}\sim N(\mathbf{0},\,\tau_\delta^2 Q_\delta^{-1}),
\end{equation}
with the type-level field $\bm\mu_{g,t}$ smoothed in space and time in the same way as the per-neuron field of Part A, and given a sparsity mechanism so that pooling neurons within a type does not reintroduce, at the type level, the fragmentation problem of Section~\ref{sec:motivation}. We state this direction as a target for future work, not as a model we have specified in full or fit: beyond the single decomposition above, nothing has been implemented, and no result anywhere in this paper depends on it. Item (i) of Section~\ref{sec:limitations} is this future work's single highest-priority piece.

\section{Conclusion}

We formulated dynamic receptive-field estimation for heterogeneous neurons as a structured, high-dimensional Bayesian inference problem: a spatio-temporally structured Poisson model -- an SPDE-based Gaussian Markov random field in space, an AR(1) process in time, fit by INLA. On $155$ salamander retinal ganglion cells, we fit this spatio-temporal component independently for each neuron. An ordinary pixel-level Poisson-LASSO fit localizes the receptive field but returns a spatially fragmented pattern with isolated, disconnected pixel selections (Section~\ref{sec:empirical-illustration}); the SPDE/AR(1) fit instead recovers a single, spatially coherent region (Section~\ref{sec:real-data}). Clustering the resulting per-neuron surfaces by their temporal response profile gives three well-balanced phenotypes ($n=85,32,38$; Section~\ref{sec:population-typing}). In a simulation study with known ground truth, the same spatio-temporal model recovers more of the true active region and estimates its magnitude more accurately than an unregularized Poisson GLM, LASSO, and the elastic net, though not uniformly: LASSO controls false positives better (Section~\ref{sec:sim-results}). Separately from all of this, and reported strictly as future work rather than as a result of this paper, Section~\ref{sec:future-work} sketches a direction for a hierarchical extension that would pool neurons of the same functional type; that extension has not been specified in full or fit.

Together, the real-data fit, the LASSO contrast, the population typing, and the simulation validation are this paper's demonstrated contribution. What is not yet demonstrated is the fully joint hierarchical fit itself: pooling neurons of the same phenotype through the shared type-level field and structured-sparsity prior of Section~\ref{sec:future-work}, and measuring what that pooling adds over the independent per-neuron fits reported here. Section~\ref{sec:limitations} states this, and the paper's other open items, in full.

More broadly, the proposed approach reframes receptive-field estimation as a structured high-dimensional Bayesian inference problem, with the potential -- once the joint fit is complete -- to provide both improved statistical recovery and a more interpretable description of visual processing across heterogeneous neural populations.

\section{Data and Code Availability}
\label{sec:data-availability}

The neural spike-train and stimulus-image data analyzed in this paper originate from the salamander retinal ganglion cell recordings described by \citet{liu2022simple}. The analysis code -- covering the pixel-level LASSO comparison (Section~\ref{sec:empirical-illustration}), the per-neuron INLA/SPDE spatio-temporal fit (Section~\ref{sec:real-data}), and the population-level clustering pipeline (Section~\ref{sec:population-typing}), including the exact figure-generating scripts -- is available from the corresponding author upon request and will be released in a public code repository alongside the completed simulation study, in enough detail to replicate all results reported here.

\bibliographystyle{abbrvnat}
\bibliography{references}

@article{zou2005,
  title={Regularization and variable selection via the elastic net},
  author={Zou, Hui and Hastie, Trevor},
  journal={Journal of the Royal Statistical Society Series B: Statistical Methodology},
  volume={67},
  number={2},
  pages={301--320},
  year={2005},
  publisher={Oxford University Press}
}

@article{freijeirogonzalez2022,
  title={A critical review of LASSO and its derivatives for variable selection under dependence among covariates},
  author={Freijeiro-Gonz{\'a}lez, Laura and Febrero-Bande, Manuel and Gonz{\'a}lez-Manteiga, Wenceslao},
  journal={International Statistical Review},
  volume={90},
  number={1},
  pages={118--145},
  year={2022},
  publisher={Wiley Online Library}
}

@article{liu2022simple,
  title={Simple model for encoding natural images by retinal ganglion cells with nonlinear spatial integration},
  author={Liu, Jian K and Karamanlis, Dimokratis and Gollisch, Tim},
  journal={PLOS Computational Biology},
  volume={18},
  number={3},
  pages={e1009925},
  year={2022},
  publisher={Public Library of Science}
}

@article{rue2017Bayesian,
  title={Bayesian computing with INLA: a review},
  author={Rue, H{\aa}vard and Riebler, Andrea and S{\o}rbye, Sigrunn H and Illian, Janine B and Simpson, Daniel P and Lindgren, Finn K},
  journal={Annual Review of Statistics and Its Application},
  volume={4},
  pages={395--421},
  year={2017},
  publisher={Annual Reviews}
}

@article{bakka2018spatial,
  title={Spatial modeling with R-INLA: A review},
  author={Bakka, Haakon and Rue, H{\aa}vard and Fuglstad, Geir-Arne and Riebler, Andrea and Bolin, David and Illian, Janine and Krainski, Elias and Simpson, Daniel and Lindgren, Finn},
  journal={Wiley Interdisciplinary Reviews: Computational Statistics},
  volume={10},
  number={6},
  pages={e1443},
  year={2018},
  publisher={Wiley Online Library}
}

@article{park2011receptive,
  title={Receptive field inference with localized priors},
  author={Park, Mijung and Pillow, Jonathan W},
  journal={PLOS Computational Biology},
  volume={7},
  number={10},
  pages={e1002219},
  year={2011},
  publisher={Public Library of Science}
}

@article{pillow2008spatio,
  title={Spatio-temporal correlations and visual signalling in a complete neuronal population},
  author={Pillow, Jonathan W and Shlens, Jonathon and Paninski, Liam and Sher, Alexander and Litke, Alan M and Chichilnisky, EJ and Simoncelli, Eero P},
  journal={Nature},
  volume={454},
  number={7207},
  pages={995--999},
  year={2008},
  publisher={Nature Publishing Group}
}

@article{rue2009approximate,
  title={Approximate {B}ayesian inference for latent {G}aussian models by using integrated nested {L}aplace approximations},
  author={Rue, H{\aa}vard and Martino, Sara and Chopin, Nicolas},
  journal={Journal of the Royal Statistical Society Series B: Statistical Methodology},
  volume={71},
  number={2},
  pages={319--392},
  year={2009},
  publisher={Oxford University Press}
}

@article{lindgren2011explicit,
  title={An explicit link between {G}aussian fields and {G}aussian {M}arkov random fields: the stochastic partial differential equation approach},
  author={Lindgren, Finn and Rue, H{\aa}vard and Lindstr{\"o}m, Johan},
  journal={Journal of the Royal Statistical Society Series B: Statistical Methodology},
  volume={73},
  number={4},
  pages={423--498},
  year={2011},
  publisher={Oxford University Press}
}

@article{jamessugar2003,
  title={Clustering for sparsely sampled functional data},
  author={James, Gareth M and Sugar, Catherine A},
  journal={Journal of the American Statistical Association},
  volume={98},
  number={462},
  pages={397--408},
  year={2003},
  publisher={Taylor \& Francis}
}

@article{dumoulinwandell2008,
  title={Population receptive field estimates in human visual cortex},
  author={Dumoulin, Serge O and Wandell, Brian A},
  journal={NeuroImage},
  volume={39},
  number={2},
  pages={647--660},
  year={2008},
  publisher={Elsevier}
}

@misc{manna2026scalablespatialpointprocess,
  title={Scalable spatial point process models for forensic footwear analysis},
  author={Manna, Alokesh and Spencer, Neil and Dey, Dipak K.},
  year={2026},
  eprint={2602.07006},
  archivePrefix={arXiv},
  primaryClass={cs.CV},
  url={https://arxiv.org/abs/2602.07006}
}

@misc{manna2026bwmp,
  title={Bigraphical {M}at\'ern-{W}hittle (BMW) Processes for Fast Inference of Big Multivariate Spatial Data on General Domains},
  author={Dey, Debangan and Manna, Alokesh and Geoga, Christopher J.},
  year={2026},
  eprint={2609.01950},
  archivePrefix={arXiv},
  url={https://arxiv.org/abs/2609.01950}
}

@misc{manna2025arlagselection,
  title={Bayesian Models for Joint Selection of Features and Auto-Regressive Lags: Theory and Applications in Environmental and Financial Forecasting},
  author={Manna, Alokesh and Ghosh, Sujit K.},
  year={2025},
  eprint={2508.10055},
  archivePrefix={arXiv},
  url={https://arxiv.org/abs/2508.10055}
}

@article{kassventura2001,
  title={A spike-train probability model},
  author={Kass, Robert E. and Ventura, Val{\'e}rie},
  journal={Neural Computation},
  volume={13},
  number={8},
  pages={1713--1720},
  year={2001},
  publisher={MIT Press}
}

@incollection{paninski2008statistical,
  title={Statistical models of spike trains},
  author={Paninski, Liam and Brown, Emery N. and Iyengar, Satish and Kass, Robert E.},
  booktitle={Stochastic Methods in Neuroscience},
  editor={Laing, Carlo and Lord, Gabriel J.},
  publisher={Oxford University Press},
  year={2008}
}

@article{reinhart2018review,
  title={A review of self-exciting spatio-temporal point processes and their applications},
  author={Reinhart, Alex},
  journal={Statistical Science},
  volume={33},
  number={3},
  pages={299--318},
  year={2018},
  publisher={Institute of Mathematical Statistics}
}

@article{davismikosch2008,
  title={Extreme value theory for space-time processes with heavy-tailed distributions},
  author={Davis, Richard A. and Mikosch, Thomas},
  journal={Stochastic Processes and their Applications},
  volume={118},
  number={4},
  pages={560--584},
  year={2008},
  publisher={Elsevier}
}

@phdthesis{manna2026dissertation,
  title={Bayesian Hierarchical Models for Complex Spatial, Temporal, and Spatio-Temporal Data: Theory, Scalability, and Applications},
  author={Manna, Alokesh},
  year={2026},
  school={University of Connecticut},
  type={Ph.D. dissertation},
  url={https://collections.ctdigitalarchive.org/node/4210739}
}

\begin{appendices}

\section{Population-level clustering: additional detail}
\label{app:clustering-detail}

This appendix records, step by step, the numerical detail behind Section~\ref{sec:population-typing} for reproducibility: how the three temporal-response phenotypes were obtained (Steps 1--4), and how Figure~\ref{fig:cluster_location_map_temporal} was then constructed from them (Steps 5--7).

\begin{enumerate}[label=\arabic*.,leftmargin=2em]
\item \emph{Per-neuron temporal profile.} For each of the $155$ neurons, average its fitted $13\times13\times30$ coefficient surface $\widehat{\bm\beta}_{k1},\ldots,\widehat{\bm\beta}_{kT}$ over the $169$ pixels at each time bin, $\bar\beta_k(t) = \tfrac{1}{169}\sum_{s=1}^{169}\widehat\beta_{kst}$ for $t=1,\ldots,30$. This gives $155$ curves, each a $30$-point vector.
\item \emph{B-spline expansion.} Project each $30$-point curve, by least squares, onto a cubic B-spline basis with $8$ basis functions. This gives $155$ vectors of length $8$, a smoothed, lower-dimensional summary of each curve's shape.
\item \emph{Functional PCA.} Run ordinary principal components analysis on the resulting $155\times8$ matrix of basis coefficients. Retain the smallest number of components, at least $2$, whose cumulative variance explained reaches $99\%$.
\item \emph{Gaussian mixture and BIC.} Fit a Gaussian mixture model (\texttt{mclust}) to the retained PCA scores, once for every candidate cluster count $k\in\{2,\ldots,10\}$ and every candidate covariance-model family; select the $(k,\text{model})$ combination with the best BIC. This selects $k=3$, with cluster sizes $85$, $32$, $38$; every neuron now carries a phenotype label $c\in\{1,2,3\}$.
\item \emph{Cluster-average spatio-temporal field.} For each phenotype $c$, return to the \emph{full} (not space-averaged) fitted surfaces of its member neurons and average them elementwise, $\overline{\bm{B}}_c = \tfrac{1}{|c|}\sum_{k\in c}\widehat{\bm\beta}_k$, a $13\times13\times30$ object -- distinct from the $30$-point curve of Step 1, since the spatial dimension is retained here.
\item \emph{Peak.} Within $\overline{\bm B}_c$, locate the single pixel-by-time-bin cell of largest absolute value, $(s^\star_c,t^\star_c) = \arg\max_{s,t}\left|\overline{B}_{c,s,t}\right|$. Pixel $s^\star_c$'s $(x,y)$ grid coordinate is the triangle plotted for phenotype $c$ in Figure~\ref{fig:cluster_location_map_temporal}; absolute value is used because the fitted coefficient can be excitatory (positive) or inhibitory (negative), and both count equally as departure from baseline for locating where the response is strongest.
\item \emph{Magnitude-weighted centroid.} Fixing time at $t^\star_c$ from Step 6, average all $169$ pixel coordinates weighted by $\left|\overline{B}_{c,s,t^\star_c}\right|$, $\text{centroid}_c = \dfrac{\sum_{s=1}^{169}(x_s,y_s)\left|\overline{B}_{c,s,t^\star_c}\right|}{\sum_{s=1}^{169}\left|\overline{B}_{c,s,t^\star_c}\right|}$. This is the circle plotted for phenotype $c$: it uses only the single time slice found in Step 6, not a value integrated across time, and applies no threshold -- every pixel contributes, in proportion to its magnitude at that time slice.
\end{enumerate}

\section{The spatial precision matrix $Q_w$: construction and the role of its hyperparameters}
\label{app:precision-detail}

This appendix gives the full algebraic detail behind the smoothing mechanism summarized in Section~\ref{sec:model-fitted}. Under the SPDE approach of \citet{lindgren2011explicit}, $Q_w$ is built directly from a finite-element discretization of a differential operator on the mesh,
\begin{equation}
Q_w = \sigma_w^{-2}\left(\kappa^4 \mathbf{C} + 2\kappa^2 \mathbf{S} + \mathbf{S}\mathbf{C}^{-1}\mathbf{S}\right),
\end{equation}
where $\mathbf{C}$ is the diagonal mass matrix (the $(m,m)$ entry is the area covered by mesh node $m$'s basis function $\psi_m$) and $\mathbf{S}$ is the stiffness matrix, $S_{ml} = \int \nabla\psi_m(\mathbf{u})\cdot\nabla\psi_l(\mathbf{u})\,d\mathbf{u}$, which is exactly zero whenever nodes $m$ and $l$ do not share a mesh triangle. $\mathbf{S}$ -- and hence $Q_w$ -- is therefore a graph-Laplacian-type operator on the mesh's adjacency structure, zero for every pair of non-neighboring nodes: this is precisely the same mechanism, a sparse precision matrix built from a graph/mesh adjacency operator, that governs the spatial coupling in our earlier point-process work on forensic footwear-impression intensity surfaces \citep{manna2026scalablespatialpointprocess}, and it is what makes $Q_w$ sparse here for the same reason it is sparse there.

This sparsity pattern is not merely a computational convenience -- it is what \emph{defines} the smoothing. By the standard Gaussian Markov random field identity, the conditional mean of one mesh-node coefficient given every other is a precision-weighted average of its neighbors alone,
\begin{equation}
E\!\left[w_{km} \mid \mathbf{w}_{k,-m}\right] = -\frac{1}{(Q_w)_{mm}} \sum_{l \sim m} (Q_w)_{ml}\, w_{kl}, \qquad (Q_w)_{ml}\ne0 \iff l\sim m,
\end{equation}
where $l\sim m$ denotes mesh nodes sharing a triangle with $m$, the same adjacency induced by $\mathbf{S}$ above: $w_{km}$ is pulled toward a weighted average of its immediate spatial neighbors and nothing else, since $(Q_w)_{ml}=0$ for every non-neighboring pair. This is the precise, algebraic mechanism by which the SPDE prior discourages the isolated, disconnected pixel selections of Figure~\ref{fig:lasso_fit}: the LASSO penalty treats every pixel's coefficient as conditionally independent of every other's given the data, so nothing pulls neighboring coefficients toward one another, whereas $Q_w$'s adjacency-structured sparsity does exactly that, by construction, for every pair of neighboring mesh nodes.

The hyperparameters $(\kappa,\sigma_w)$ control two distinct aspects of this coupling rather than one. $\kappa$ sets the spatial range -- reparametrized in Section~\ref{sec:inla-inference} to a practical correlation range, the distance at which correlation drops to about $0.1$ -- and enters $Q_w$ nonlinearly: as $\kappa$ grows, the $\kappa^4\mathbf{C}$ term (purely diagonal, purely local) comes to dominate the other two terms, which scale more slowly in $\kappa$, so larger $\kappa$ pushes $Q_w$ toward a more diagonal, more local matrix -- weaker relative coupling between neighbors and a shorter correlation range -- while smaller $\kappa$ leaves the neighbor-coupling terms $\mathbf{S}$ and $\mathbf{S}\mathbf{C}^{-1}\mathbf{S}$ relatively larger, producing stronger coupling and a longer-range, smoother field. $\sigma_w$ instead rescales $Q_w$ by a single multiplicative constant ($\sigma_w^{-2}$ above), so it governs the overall marginal variance of $w_k(\mathbf{u})$ without changing the relative pattern of coupling between neighbors at all. Both are given penalized-complexity priors, not fixed at arbitrary values (Section~\ref{sec:inla-inference}), so it is the data -- not an ad hoc smoothing choice -- that determines how strongly nearby pixels are tied together in the fitted field.

\end{appendices}

\end{document}